\documentclass[conference]{IEEEtran}

\AtBeginDocument{%
  \providecommand\BibTeX{{%
    Bib\TeX}}}

\IEEEoverridecommandlockouts
\usepackage{cite}
\usepackage{amsmath,amssymb,amsfonts}
\usepackage[ruled,vlined]{algorithm2e}
\usepackage{algorithmic}
\usepackage{graphicx}
\usepackage{textcomp}
\usepackage{xcolor}
\usepackage{multirow}
\usepackage{hyperref}
\usepackage{graphicx}
\usepackage{subcaption}
\def\BibTeX{{\rm B\kern-.05em{\sc i\kern-.025em b}\kern-.08em
    T\kern-.1667em\lower.7ex\hbox{E}\kern-.125emX}}

\newcommand{\includegraphicsorblank}[2][\linewidth]{%
    \IfFileExists{#2}{%
        \includegraphics[#1]{#2}%
    }{%
        \fbox{\parbox[c][5cm][c]{#1}{\centering \textbf{Missing Image}}}%
    }%
}

\newcommand{\ie}{\emph{i.e.} }

\newcommand{\eg}{\emph{e.g.} }

\begin{document}

\title{AdaScale: Dynamic Context-aware DNN Scaling via Automated Adaptation Loop on Mobile Devices\\

}


\author{Yuzhan Wang, 
Sicong Liu*,~\IEEEmembership{Member,~IEEE,}
Bin Guo*,~\IEEEmembership{Senior Member,~IEEE,}\\
Boqi Zhang,
Ke Ma,
Yasan Ding,
Hao Luo,
Yao Li,
Zhiwen Yu,~\IEEEmembership{Senior Member,~IEEE,}
\thanks{{*}\textit{Corresponding author: Bin Guo (e-mail:guob@nwpu.edu.cn), Sicong Liu (e-mail:scliu@nwpu.edu.cn})

Yuzhan Wang, Sicong Liu, Bin Guo, Boqi Zhang, Ke Ma, Yasan Ding, Hao Luo, Yao Li, and Zhiwen Yu were with the Department of Computer Science, Northwestern Polytechnical University, Xi'an 710072, China (e-mail: guob@nwpu.edu.cn, scliu@nwpu.edu.cn)}
}


\maketitle

\begin{abstract}
Deep learning is reshaping mobile applications, with a growing trend of deploying deep neural networks (DNNs) directly to mobile and embedded devices to address real-time performance and privacy. 
To accommodate local resource limitations, techniques like weight compression, convolution decomposition, and specialized layer architectures have been developed.
However, the \textit{dynamic} and \textit{diverse} deployment contexts of mobile devices pose significant challenges.
Adapting deep models to meet varied device-specific requirements for latency, accuracy, memory, and energy is labor-intensive. 
Additionally, changing processor states, fluctuating memory availability, and competing processes frequently necessitate model re-compression to preserve user experience.
To address these issues, we introduce AdaScale, an elastic inference framework that automates the adaptation of deep models to dynamic contexts. 
AdaScale leverages a self-evolutionary model to streamline network creation, employs diverse compression operator combinations to reduce the search space and improve outcomes, and integrates a resource availability awareness block and performance profilers to establish an automated adaptation loop.
Our experiments demonstrate that AdaScale significantly enhances accuracy by 5.09\%, reduces training overhead by 66.89\%, speeds up inference latency by 1.51 to 6.2 $\times$ , and lowers energy costs by 4.69 $\times$.


\end{abstract}


\section{Introduction}
\label{Introduction}


Deep learning has significantly advanced mobile applications such as automated driving assistance on Roadside cameras~\cite{cho2022autonomous,pechinger2023roadside,liu2023enabling}, video surveillance on UAV/UAG~\cite{rosser2018surgical}, biometric authentication on smartphones~\cite{mahfouz2017survey}, and emotion detection on smartphones~\cite{zhang2018moodexplorer, wang2023attrleaks, chu2023nnperf}.
With demands for near-/real-time performance and privacy, deploying deep models to mobile and embedded devices~\cite{chen2020deep,liu2021adaspring} has become a trend, enabling local data processing without cloud transmission.
Various handcrafted and on-demand deep model compression techniques
have been proposed, including weight compression~\cite{ko2017adaptive,dettmers20218},
convolution decomposition~\cite{tan2019mixconv,fang2024adashadow}, and special layer architectures~\cite{yu2024inceptionnext,chen2020dynamic,zhang2024deep}, to fit local resource constraints.

However, mobile devices typically operate in \textit{dynamic} and \textit{diverse} deployment environments. 
For instance, a text-based emotion detection app utilizing a BERT model could be installed on millions of smartphones/wearables, spanning from low- (\eg CPU) to high-end devices (\eg GPU, NPU) with resource availabilities that vary up to 20$\times$ and fluctuate over time. Specifying deep models for each device type and hardware context to satisfy demands on latency, accuracy, memory, or energy cost is labor-intensive~\cite{liu2024adaknife, wang2021context}. 
Similarly, multimodal recognition models in different cockpit systems may face varied and changing computational capacities under strict latency requirements for safety. Resource availability evolves due to processor states, memory availability, and competing processes. 
For example, loading parameters or activations from L2-cache is $20$ $\times$ faster than from DRAM, but cache-hit-rates change over time due to factors like operator size and cache availability. 
To maintain a stable user experience, developers often need to \textit{re-compress} deep models to adapt to dynamic contexts.

Given this challenge, a variety of prior efforts for specifying deep models suited to particular environments. 
These techniques generally fall into two categories: the \textit{pre-deployment} method~\cite{cai2019once,cai2018proxylessnas,liu2018progressive,liu2018darts}, where models are designed and finalized in the cloud before being sent to mobile devices, and the \textit{post-deployment} method~\cite{liu2021adaspring,wen2023adaptivenet,han2021legodnn,hou2022neulens}, which allows models to adjust to the dynamic context after being deployed. 
The latter scheme is often more cost-effective and preferable for mobile devices with diversity and dynamics, offering a novel approach that enables a more suitable model architecture within the target deployment context and enhances user privacy by avoiding the collection of user data.
Specifically, methods like AdaSpring~\cite{liu2021adaspring}, AdaptiveNet~\cite{wen2023adaptivenet}, and LegoDNN~\cite{han2021legodnn} exemplify this trend by producing DNNs with adaptable structures to meet the specific demands of edge devices' runtime environments, thus optimizing accuracy within given resource and latency limits. 
These dynamically adaptive methods have significantly advanced the deployment of deep models on mobile devices. Despite these advances, developers continue to encounter the following challenges:

\begin{figure*}[t]
\centerline{\includegraphics[width=1\textwidth]{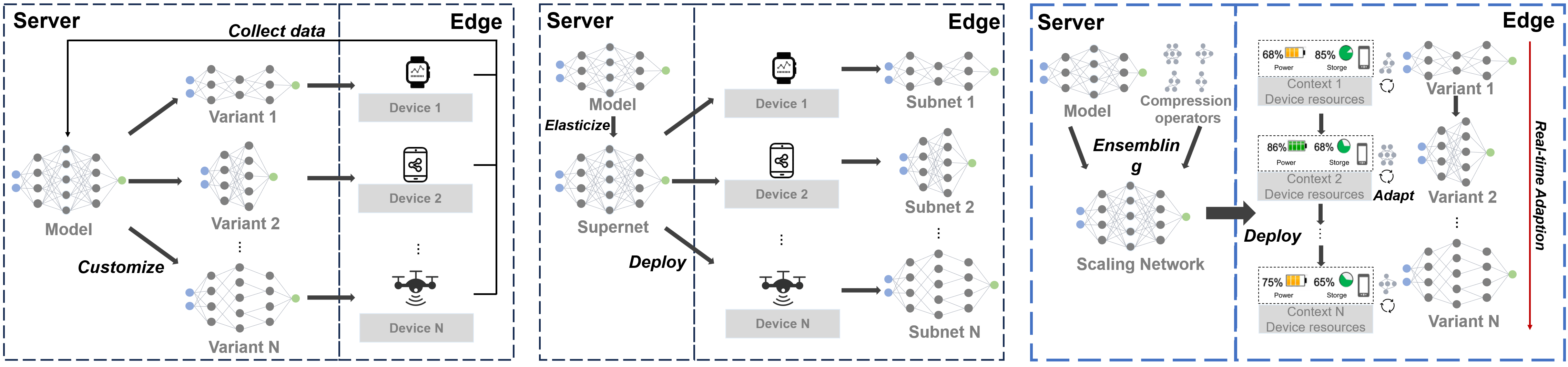}}
\caption{Comparison of different model adaptive deployment approaches. Left: pre-deployment on-server model generation. Middle: post-deployment on-device model adaptation. Right: \textbf{real-time post-deployment on-device model adaptation} (ours).}
\label{fig_deploy_method}
\end{figure*}

\begin{itemize}
\item \textit{Lightweight Scalable Model}. Existing methods often modify large networks like ResNet and VGG into reusable blocks to build scalable DNNs for various contexts~\cite{}. 
However, these models are not inherently suited for mobile devices, leading to a surplus of discarded structures and a vast search space. 
Using lighter-weight scalable models is crucial to align model complexity with device resource limitations.


\item \textit{Adaptation Frequency}. Prior scalable DNNs typically operate on a "Once for all" basis~\cite{cai2019once}. Frequent variant switching can introduce additional operational latency and limit the effective use of computational resources. A suitable trigger strategy for adaptation is necessary.

\item \textit{Runtime Profiler}. 
Scalable methods for deploying DNNs typically focus on DNN-centric metrics, such as computational load and accuracy, and struggle with profiling hardware-dependent performance indicators like latency and energy consumption~\cite{shantharama2020hardware}.
This results in a sub-optimal evaluation of adaptation strategies. 
A timely and accurate resource and performance profiler is desired.

\end{itemize}

Given these challenges, we propose AdaScale, a multi-granularity, operator-ensembled, elastic inference framework designed to automate the dynamic adaptation of deep models to varying contexts. 
AdaScale leverages a pre-trained self-evolutionary model to facilitate this automation, significantly streamlining the creation of network structures by employing diverse compression operator combinations. This not only reduces the search space but also improves outcomes compared to random network generation.
AdaScale features a multi-branch early exit structure to enhance dynamic resource adaptation and utilizes a multi-stage training mode that optimizes network performance while reducing training costs, boosting network efficiency and environmental sustainability. Additionally, it integrates blocks for resource availability awareness and performance profiling to precisely initiate adaptations, ensuring that models capitalize on real-time device capabilities and enhance overall adaptability.
The main contributions of this work are as follows:

\begin{itemize}
    \item To the best of our knowledge, AdaScale represents the first work that integrates device resource awareness with model performance optimization into a context-adaptive deep neural network inference framework. It utilizes lightweight network blocks and multi-branch strategies to construct a self-evolutionary elastic network that continuously scales and adapts. Moreover, AdaScale incorporates a multi-stage training approach complemented by a multi-branch exit mechanism, significantly enhancing the network's adaptability to real-time changes.
    
    \item AdaScale proposes an efficient runtime search strategy to effectively address the challenge of runtime resource adaptation. It features a flexible combination of lightweight compression operators, rapid resource sensing, search strategies, and a robust training mechanism that significantly enhances the network’s adaptability, responsiveness, and performance during runtime.
    \item Experiments show that the DNNs generated by AdaScale achieve comparable performance to existing handcrafted compression techniques under various user demands. Through extensive cross-platform experimentation involving multiple tasks and various advanced adaptive and lightweight methodologies, AdaScale has demonstrated substantial benefits in deploying DNNs under resource-constrained conditions.    It significantly enhances accuracy by 5.09\%, reduces training overhead by 66.89\%, speeds up inference latency by 1.51 to 6.2 $\times$ , and lowers energy costs by 4.69 $\times$, all while maintaining less than 4\% accuracy loss in resource-deficient environments..

\end{itemize}

In the rest of this paper, we present the system overview in $\S$~\ref{OVERVIEW}, and elaborate the design in $\S$~\ref{COMPRESSION OPERATOR-ENSEMBLED SELF-EVOLUTIONARY NETWORK} and $\S$~\ref{RUNTIME ELASTIC MODEL ADJUSTMENT}. 
And then we review the related work in $\S$~\ref{RUNTIME ELASTIC MODEL ADJUSTMENT}, evaluate AdaScale in $\S$~\ref{EVALUATION}, and conclude in $\S$~\ref{CONCLUSION}.

\section{OVERVIEW}
\label{OVERVIEW}
This section starts with problem analysis and then presents an overview of AdaScale design.

\subsection{Problem Study}
\label{Problem Study}



As shown in Figure~\ref{fig_deploy_method}, we analyze three deployment methods for DNNs on mobile devices. The first method generates a sub-model in the cloud and deploys it to the device, requiring substantial data uploads and potentially compromising user privacy. The second method is on-device adaptation with a static network configuration, lacking flexibility to adapt to changes in device context. The third, our proposed method, introduces a dynamic deployment model where the DNN evolves in real-time, adapting to the device's changing context and offering a flexible, context-aware solution.

Our approach focuses on continuously adapting DNN structures based on device context to optimize performance metrics such as latency, energy consumption, and accuracy. We have conducted extensive testing of various deep learning models across different contextual scenarios to evaluate their performance. Additionally, we have analyzed the correlation between DNN operations and device runtime states for more accurate and efficient modeling. This enables our DNNs to operate continuously in mobile environments, meeting stringent requirements for efficiency and effectiveness, thereby presenting a significant advancement in the dynamic deployment of neural networks on mobile devices.


\begin{itemize}
    \item \textbf{Accuracy} is crucial for high-quality task completion with DNNs. Efficient training of networks of varying sizes is essential to uphold performance standards.
    \item \textbf{Responsiveness} requires that DNN needs to dynamically adapt to meet varying task requirements, especially with limited computing resources.
    \item \textbf{Energy efficiency} is critical for the sustainable operation of DNNs on sensor platforms. Continually enhancing energy efficiency is vital to resolving a primary bottleneck in mobile computing.
    \item \textbf{Dynamic adaptivity} requires DNNs to be hardware-aware and capable of dynamically adjusting their architecture and performance in response to different contexts. This adaptability necessitates real-time monitoring of device performance and optimization to balance response time, energy efficiency, and accuracy, thereby meeting the evolving needs of mobile deployments.
    \item \textbf{Dynamic adaptivity} requires DNNs to be hardware-aware and dynamically adjust their architecture and performance based on contextual conditions. This adaptability necessitates real-time monitoring of device performance and optimization to balance response time, energy efficiency, and accuracy, meeting the evolving needs of mobile deployments.
\end{itemize}

Despite previous efforts discussed in $\S$~\ref{RELATED WORK}, current solutions have not fully met the demands for effectively deploying DNNs on mobile devices. To overcome these limitations, this paper introduces AdaScale, a resource-aware multi-variant elastic scaling inference framework. AdaScale is designed to optimize the interaction between mobile device capabilities and DNN requirements for continuous deep learning tasks, optimizing both performance and resource utilization.


\subsection{Problem Formulation}
\label{Problem Formulation}



\begin{figure*}[t]
\centerline{\includegraphics[width=0.95\textwidth]{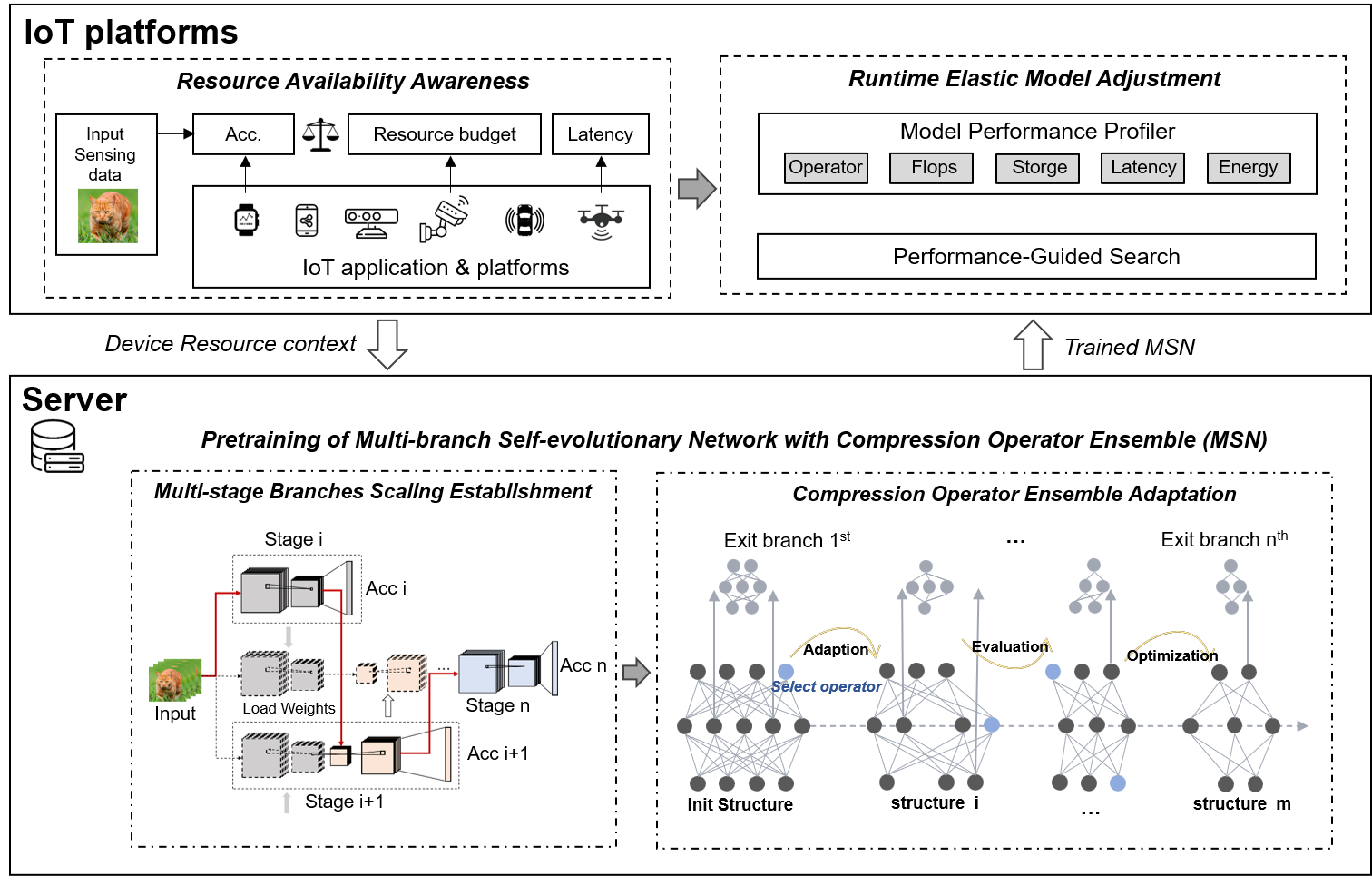}}
\caption{The framework of AdaScale includes two main components: the pretraining of a multi-branch self-evolutionary network with a compression operator ensemble on the server , and the awareness of resource availability with runtime elastic model adjustment on the IoT device.}
\label{fig_framework}
\end{figure*}

As depicted in Figure~\ref{fig_framework}, AdaScale utilizes a comprehensive strategy to balance deep neural network (DNN) performance by integrating multiple compression techniques. This approach targets the optimization of interrelated performance metrics, ensuring that the most effective model is selected for mobile devices with fluctuating computational resources. We tackle the issue of deploying deep learning models on mobile devices with constrained computational power by establishing a dynamic optimization framework. The primary objective is to reduce the discrepancy between a model's performance demands and a device's processing abilities, while also meeting strict accuracy and resource constraints. The dynamic optimization problem addressed by AdaScale is formulated as:

\begin{equation} \label{eq:mm}
\begin{aligned}
& \min_{P} \sqrt{\sum_{i}^{n} \sum_{j}^{m} (P[i] - C_{d}[j])^2}, \\
& \text{s.t. }Acc \geq Acc_{u}, avgP \leq avgC_{d}
\end{aligned}
\end{equation}
where the model's performance requirements (\( P \)) encompass vectors for energy consumption (\( E \)), latency (\( L \)), accuracy (\( Acc \)), and response time (\( R \)): \( P = [E, L, Acc, R] \). Device computational capabilities (\( C_{d} \)) include the device's CPU (\( C_{cpu} \)), GPU (\( C_{gpu} \)), and memory (\( C_{mem} \)) resources: \( C_{d} = [C_{cpu}, C_{gpu}, C_{mem}] \). The optimization aims to minimize the Euclidean distance between \( P \) and \( C_{d} \), ensuring that the model's demands align closely with the device's capabilities. $\S$~\ref{Compression Operator Ensembling} will provide a detailed introduction to the ensemble of compression operators. $\S$~\ref{Multi-branch Self-evolutionary Network Pretraining} will discuss the multi-stage branch pretraining approach. $\S$~\ref{Resource Availability Awareness} will explore the dynamic device resource awareness module, while $\S$~\ref{Performance-Guided Search} will focus on the model performance awareness adaptation module. Additional details on runtime adjustments and network elasticity will be elaborated in $\S$~\ref{RUNTIME ELASTIC MODEL ADJUSTMENT}.


\subsection{Challenges}
\label{Challenges}


Deploying DNNs on mobile devices introduces several significant challenges:

Challenge \#1: \textit{Reducing redundancy in model spaces.}
Standard architectures like ResNet~\cite{he2016deep} and VGG~\cite{simonyan2014very} utilize reusable code blocks, which contribute to a large and redundant search space, inefficient for edge computing. Reducing generation costs and the search space demands redefining the model generation subspace, a task complicated by the diversity and complexity of current DNNs.

Challenge \#2: \textit{Dynamic adaptation under varying contexts.} Traditional DNN methods usually follow a once-for-all strategy, resulting in prolonged adaptation, increased latency, and inefficient resource use under dynamic conditions. Developing networks that rapidly adapt to the variable contexts of mobile devices is essential yet challenging.

Challenge \#3: \textit{Managing device performance metrics.} While research has optimized DNNs for accuracy, latency, and computational, key metrics like memory, power, and energy consumption often remain unaddressed. Real-time monitoring introduces substantial overheads. Effectively managing these metrics in adaptive operations, especially with the variable performance of mobile devices, is a critical challenge.

Each of these challenges represents a barrier to efficient DNN deployment on mobile devices, and overcoming them is crucial for advancing mobile computational capabilities.

\subsection{AdaScale Framework}
\label{AdaScale Framework}


As shown in Figure~\ref{fig_framework}, AdaScale combines several key components to overcome critical deployment challenges:
\begin{itemize}
    \item \textit{Multi-variant scaling network block} is a core component of the AdaScale Framework, which enhances model efficiency by ensembling lightweight compression operators. This design reduces the candidate space during training and generation, allowing the network to swiftly adjust to the variable runtime resource contexts of mobile devices. Its multi-stage parameter sharing mechanism ensures effective adaptation without the retraining. 
    \item \textit{Runtime elastic model adjustment block} uses a multi-branch early exit mechanism and performance search strategy, adapting quickly to contexts, maximizing resources, and ensuring optimal deep learning performance.
    \item \textit{Resource availability awareness block} monitors and evaluates the resources available on mobile devices and the performance of DNNs in real-time. It provides accurate, dynamic assessments and facilitates the scaling of network variants, offering deep insights into both device capabilities and DNN efficiency.

\end{itemize}

\section{COMPRESSION OPERATOR-ENSEMBLED SELF-EVOLUTIONARY NETWORK}
\label{COMPRESSION OPERATOR-ENSEMBLED SELF-EVOLUTIONARY NETWORK}

This section details the design of the compression operator ensemble and the multi-branch self-evolutionary network. The ensemble of compression operators is composed of lightweight blocks from various lightweight networks. Concurrently, the multi-branch self-evolutionary network integrates a high-performance backbone network with multiple evolutionary branches, which allows for dynamic adaptation and scalability.

\subsection{Multi-variant Network Architecture Ensemble}
\label{Multi-variant Network Architecture Ensembling}

This section details the construction of variant networks, including the integration of compression operators and the establishment of multi-branch networks, to enhance performance in resource-constrained environments.

\subsubsection{Compression Operator Ensemble} 
\label{Compression Operator Ensembling}



In contrast to existing methods such as LegoDNN~\cite{han2021legodnn} and Adaptivenet~\cite{wen2023adaptivenet}, which generate vast and expensive search spaces encompassing up to \(1.68 \times 10^6\) and \(2.57 \times 10^{17}\) configurations respectively. To address the problem of large search spaces and high model generation and training costs, we propose a lightweight operator integration method, offering a new approach to model generation. This method uses lightweight compression operators, proven effective in leading architectures. By integrating these operators, we simplify network construction and reduce the complexity and cost during the search and training phases. To maximize efficiency and maintain or enhance performance, our framework systematically categorizes and tailors compression operators to optimize specific aspects of neural network structures and operations.

\begin{itemize}
\item Structural optimization operator integrates methods like depthwise separable convolutions, compact layer architectures, and optimized kernel allocation to decrease computational costs and simplify the convolutional framework. It enhances resource efficiency and streamlines connections within neural networks~\cite{howard2017mobilenets,iandola2016squeezenet,zhang2018shufflenet,szegedy2015going,
szegedy2016rethinking,han2020ghostnet}.
\item Automated machine learning and NAS operator dynamically balances aspects such as network width, depth, and resolution. It leverages neural architecture search to automatically identify and configure optimal structures, significantly reducing manual intervention and potential bias in model design~\cite{tan2019efficientnet,tan2019mnasnet}.
\item Advanced connectivity operator implements dense connections to bolster information flow and gradient propagation across layers. This enhancement boosts the learning efficiency and robustness of neural networks~\cite{iandola2014densenet}.
\item Mathematical decomposition techniques employ batch normalization and decomposed convolutions to minimize model size and complexity, enabling deeper compression without sacrificing performance. Singular value decomposition further enhances efficiency by streamlining the network’s weight matrix~\cite{szegedy2016rethinking,wu2018deep}.
\end{itemize}

\subsubsection{Multi-branch Establishment}
\label{Multi-branch Establishment}

Current dynamic adaptation methods like Adaspring~\cite{liu2021adaspring} and Adadeep~\cite{liu2020adadeep} only allow pre/post-deployment modifications and are too slow to unsuitable for real-time applications. Similarly, early exit strategies such as NeuLens~\cite{hou2022neulens} and AIMA~\cite{huang2022enabling} fail to adapt during runtime. As a result, there is a notable deficiency in mechanisms that can dynamically adjust DNNs in real-time to accommodate fluctuations in device resources.

To address the limitations, we introduce a novel multi-branch, self-evolving network architecture designed for real-time adaptability to computational fluctuations. This architecture merges a robust backbone network with variant branches, each employing unique convolutional compression strategies. Our multi-branch design supports varied performance demands and leverages multi-stage training with shared parameters, drastically cutting down the frequency of weight retraining and expediting inference speeds. This adjustment ensures that the network's performance consistently matches the available computational resources.

The backbone of our network is built for flexibility, facilitating the seamless interchange of compression operators to scale with varying demands. It features a dynamic path selection mechanism at each layer, optimizing computational efforts according to task specifics. Additionally, an integrated early exit strategy allows for processing termination at intermediate stages based on input complexity and classification confidence, boosting responsiveness and efficiency in mobile environments. Current architectures like Candidate DNNs, Intermediate Classifiers, and Multi-scale Architectures offer the basis for multi-branch early exits~\cite{teerapittayanon2016branchynet,han2015learning}, capitalizing on the varying complexity of real-world inputs. While effective for basic applications, these architectures falter in complex scenarios typical of intelligent IoT devices, often prioritizing accuracy at the expense of operational speed, which results in unacceptable latency for mobile applications.

Recognizing the importance of resource for operational continuity in dynamic environments, we propose an early-exit network architecture for resource-constrained settings. This multi-branch approach dynamically adapts using early-exit blocks to terminate processing when outcomes are sufficiently determined, reducing computational load and latency.

Our early-exit blocks features a 2D convolutional layer that efficiently extracts essential features using a 3x3 convolutional kernel with a stride of 2, significantly reducing data volume. This is followed by adaptive average pooling that further reduces dimensionality to a 1x1 format, cutting down the number of parameters and streamlining the data for classification. To mitigate the risk of overfitting, a Dropout layer is incorporated, succeeded by a sparse fully connected layer that handles classification tasks without excessive resource use. 


To enhance scalability and manage resources efficiently in multi-branch early-exit networks, we introduce a testing methodology. This method integrates a simplified exit branch at every layer, tailored to adapt to varied terminal resources without the substantial overhead commonly associated with extensive branching. This overhead often impairs performance and enlarges the search space for optimal configurations. 

Our approach incorporates streamlined exit branches, each consisting of pooling and fully connected layers, enabling precise evaluation of performance distribution throughout the DNN. This strategic configuration reduces training expenses, accelerates the testing cycle, and mitigates performance skew typically introduced by downsampling layers during evaluations. Our method efficiently balances branching complexity and resource limitations, offering a scalable solution.

In the Establishment of scalable networks with multiple branches, we identified key phenomena referred to as ``branch diversity establishment'', highlighting differences in performance outcomes and enhancements among various branches. This encompasses:

\begin{itemize}
    \item \textit{Performance differences among branches:} within a multi-branch scaling network structured around backbone architectures like ResNet, significant variations in performance across structurally repeated layers were observed. Deeper layers often exhibit enhanced performance later in the training cycle, whereas shallower layers consistently underperform, leading us to strategically integrate candidate branches after high-performing layers.
    \item \textit{Performance improvement differences between neighboring branches:} there is a pronounced disparity in performance improvement between adjacent layers. Some layers demonstrate significant accuracy improvements early in training, marking them as crucial, while others show minimal enhancements even after extended training periods and are deemed less critical. Identifying these less critical layers has allowed us to target them for the application of compression operators, optimizing network efficiency while maintaining robust performance, crucial for meeting the dynamic demands of mobile computing.
\end{itemize}

These observations guide the strategic placement and optimization of branches within the network, ensuring efficient operation and adaptation to diverse computing environments.

\subsection{Multi-branch Self-evolutionary Network Pretraining}
\label{Multi-branch Self-evolutionary Network Pretraining}

\begin{figure}[t]
\centerline{\includegraphics[width=0.48\textwidth]{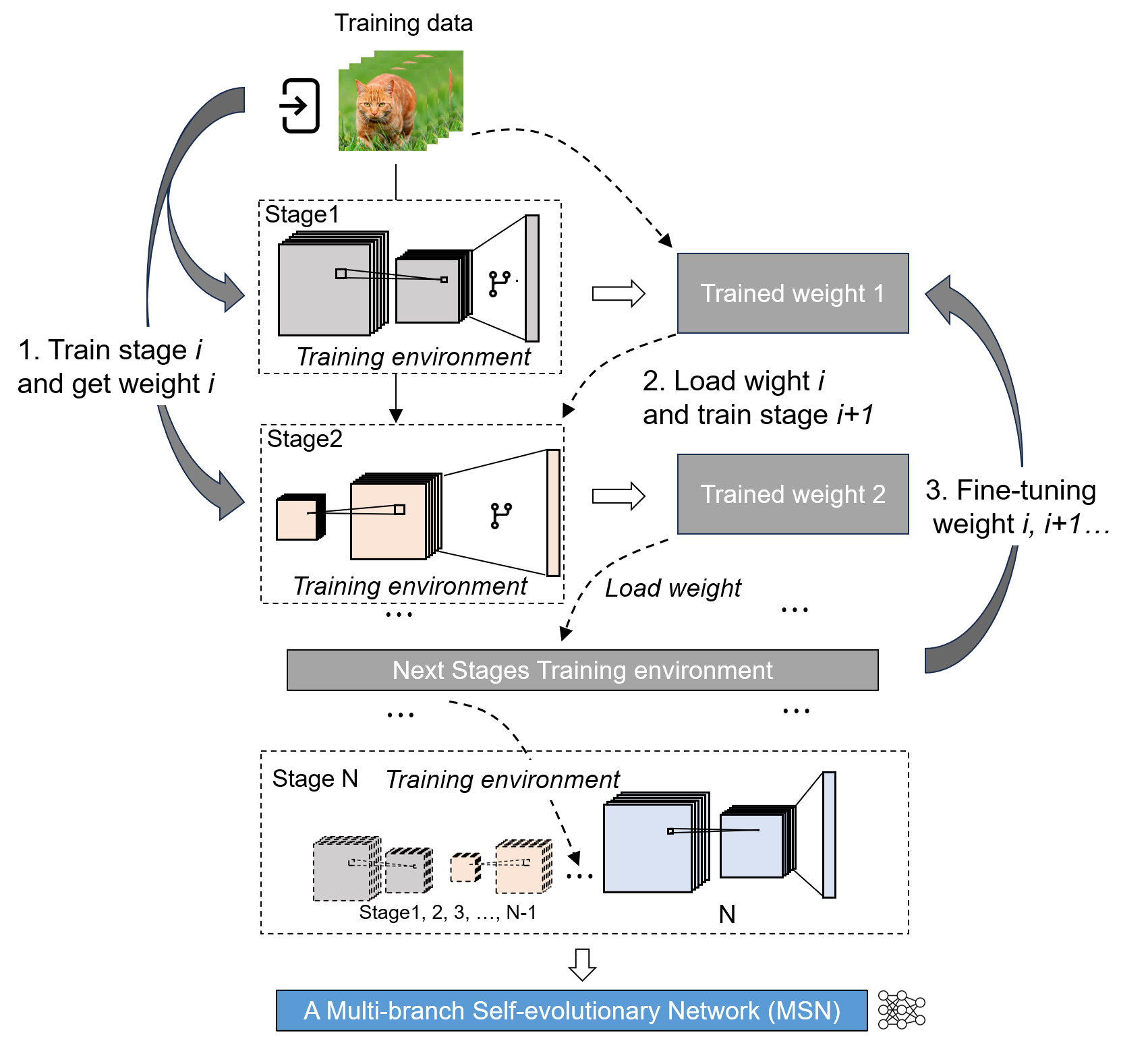}}
\caption{Illustration of multi-branch self-evolutionary network.}
\label{fig_stage-training}
\end{figure}


Multi-branch self-evolutionary network aim to deliver precise classification across multiple exit branches. Existing methodologies like Once-for-All and AdaSpring, while optimizing for high accuracy at the final classification stage, often overlook the performance at earlier exit points. This oversight not only results in poorer early classifications but also escalates time and resource expenditure. 

Inspired by findings from Wei et al. \cite{wei2023nn}, which suggest that networks of similar scale can achieve comparable accuracy, we propose a novel multi-stage training strategy. This approach not only targets enhancing precision at each exit but also focuses on reducing overall training costs. By incrementally adjusting model complexity and refining each branch, our method ensures balanced optimization of the entire network, promoting efficient resource utilization and robust performance across all branches.



Figure~\ref{fig_stage-training} illustrates how the deep neural network (DNN) is structured by segmenting it into distinct parts at designated branch exits, resulting in partitioned networks termed \(\text{Partnet}_i\) (\(p_{i}\)). Specifically, the network is divided into segments ranging from \(\text{input} \sim \text{exit}_1\), \(\text{exit}_1+1 \sim \text{exit}_2\), to \(\text{exit}_{n-1}+1 \sim \text{final}\), where \(n\) denotes the number of partitions, thus forming the complete network: \(\text{Net} = \text{$p$}_1 \cup \text{$p$}_2 \cup \ldots \cup \text{$p$}_n\).

Our training approach, detailed in Algorithm~\ref{alg:partition_training}, starts with the training of \(\text{$p$}_1\) until its accuracy at \(\text{exit}_1\) (\(\text{Acc}_1\)) meets predefined criteria, at which point \(\text{$p$}_1\) training is halted. Training for \(\text{$p$}_2\) leverages both \(\text{Acc}_1\) and the weights from the first segment (\(\text{weight}_1\) or \(w_1\)), taking advantage of the shared core components between \(\text{$p$}_1\) and \(\text{$p$}_2\) (excluding exit-specific downsampling and fully connected layers). This approach allows the direct reuse of \(w_1\), ensuring parameter continuity and exploiting pre-trained weights.

We employed two weight updating methods: 
(a) Fixing \(\text{$w$}_1\) to prevent changes, while updates are focused solely on \(\text{$w$}_2\) within \(\text{$p$}_2\). This parameter freezing technique, novel in its application to training subsequent network stages, is akin to practices used in network pruning and quantization. 
(b) Updating \(\text{$w$}_1\) if \(\text{Acc}_{1_{\text{new}}} > \text{Acc}_1\), which allows for rapid progression in training \(\text{$p$}_2\) while fine-tuning \(\text{$p$}_1\). This method is viable for networks with fewer branches, where updating costs are manageable. The process continues in a similar fashion through \(\text{$p$}_3\) and beyond, based on the accumulated accuracies and weights from preceding segments, repeating until all segments are optimally trained.

\begin{algorithm}[t]
\caption{Multi-branch Self-evolutionary Network Pretraining Algorithm}
\label{alg:partition_training}
\KwIn{A multi-branch network with \(n\) branches, 
        \(\text{$w$}_i\) for $p_i$, 
        \(\text{Acc}_i\) for each branch thresholds}
\For{$i \gets 1$ \KwTo $n$}{
    \eIf{$i == 1$}{
        Train \(\text{$p$}_1\) achieves \(\text{Acc}_1\) or meets requirements\;
        Save \(\text{$w$}_1\) and \(\text{Acc}_1\)\;
    }{
        Load \(\text{$w$}_{i-1}\) into \(\text{$p$}_{i-1}\)\;
        Train \(\text{$p$}_i\);
        \eIf{(a)}{
            Fix \(\text{$w$}_{i-1}\) and update only \(\text{$w$}_i\) through backpropagation\;
        }{
            Update \(\text{$w$}_{i-1}\) only if \(\text{Acc}_{i-1\_new} > \text{Acc}_{i-1}\)\;
        }
        Continue training \(\text{$p$}_i\) until it achieves the \(\text{Acc}_i\) or meets user-defined training requirements\;
        Save \(\text{$w$}_i\) and accuracy \(\text{Acc}_i\)\;
    }
}
\KwOut{The complete network \(\text{Net}\), with all branch networks \(\text{$p$}_1, \text{$p$}_2, \ldots, \text{$p$}_n\) and  \(\text{$w$}_1, \text{$w$}_2, \ldots, \text{$w$}_n\)  trained to the desired accuracies}
\end{algorithm}


We employ Algorithm~\ref{alg:partition_training} to thoroughly train the multi-branch self-evolving network, utilizing predefined branch positions that are determined during the network's branching construction. This training approach ensures each branch achieves precise accuracy, yielding a network that is both elastic and scalable. Such an architecture is particularly beneficial across a variety of mobile device scenarios, accommodating dynamic resource availability and computational demands. Further details on branch selection, compression operator configuration, and dynamic adaptive deployment are discussed in $\S$~\ref{Performance-Guided Search}, where we detail how to optimize network performance tailored to specific operational environments.

\section{RUNTIME ELASTIC MODEL ADJUSTMENT}
\label{RUNTIME ELASTIC MODEL ADJUSTMENT}




This section describes how AdaScale accurately predicts the computational resources of intelligent mobile devices. AdaScale then instructs the scalable network to select the most suitable operators from a pool of pre-trained compression blocks. This mechanism allows the network to dynamically adjust to the fluctuating resource capacities of devices, ensuring efficient, real-time performance customization.

\subsection{Resource Availability Awareness}
\label{Resource Availability Awareness}



Deploying DNNs on mobile devices involves managing latency and adjusting model scales to match device capabilities, as demonstrated by AdaptiveNet~\cite{wen2023adaptivenet}, AdaSpring~\cite{liu2021adaspring}, and AdaDeep~\cite{liu2020adadeep}. However, these methods predominantly address latency and model scaling, neglecting other critical computational resources such as memory, energy consumption, and storage, which are essential for effective deep learning deployment on mobile devices. This narrow focus limits their applicability in optimizing the full range of device capabilities.

To overcome the limitations of deploying DNNs on mobile devices, we propose a method for resource management, \ie \textit{resource awareness module}. The resource awareness module continuously monitors device resources and generates a device load index. Leveraging an autoregressive model, it forecasts short-term future loads, enabling proactive adaptation to the stringent computational, storage, and energy constraints characteristic of dynamic mobile device environments.

Our method involves real-time monitoring of critical shared resources, including memory, CPU, and GPU. We focus on collecting data on utilization rates and operating frequencies. To enhance computational load analysis, we introduce a device load index, which integrates usage levels of the CPU, GPU, and memory resources. This index considers the usage levels of the CPU, GPU, and memory resources. The formula for the load index is defined as follows:

\begin{equation} \label{eq:load_index}
I_{\text{load}} = (W_{cpu} \times U_{cpu} + W_{gpu} \times U_{gpu} + W_{M} \times U_{M}) / W,
\end{equation}
where \( I_{\text{load}} \) represents the load index, \( U_{cpu} \), \( U_{gpu} \), and \( U_{M} \) denote the utilization of the CPU, GPU, and memory, respectively. The coefficients \( W_{cpu} \), \( W_{gpu} \), and \( W_{M} \) are the respective weights assigned to the CPU, GPU, and memory, with \( W \) being the sum of these weights. The monitoring of the CPU, for example, involves calculating the difference between system state snapshots taken from the \textit{/proc/stat} file at two distinct times (\textit{prev} and \textit{current}) to measure CPU usage.

\begin{figure*}[t]
\centerline{\includegraphics[width=0.95\textwidth]{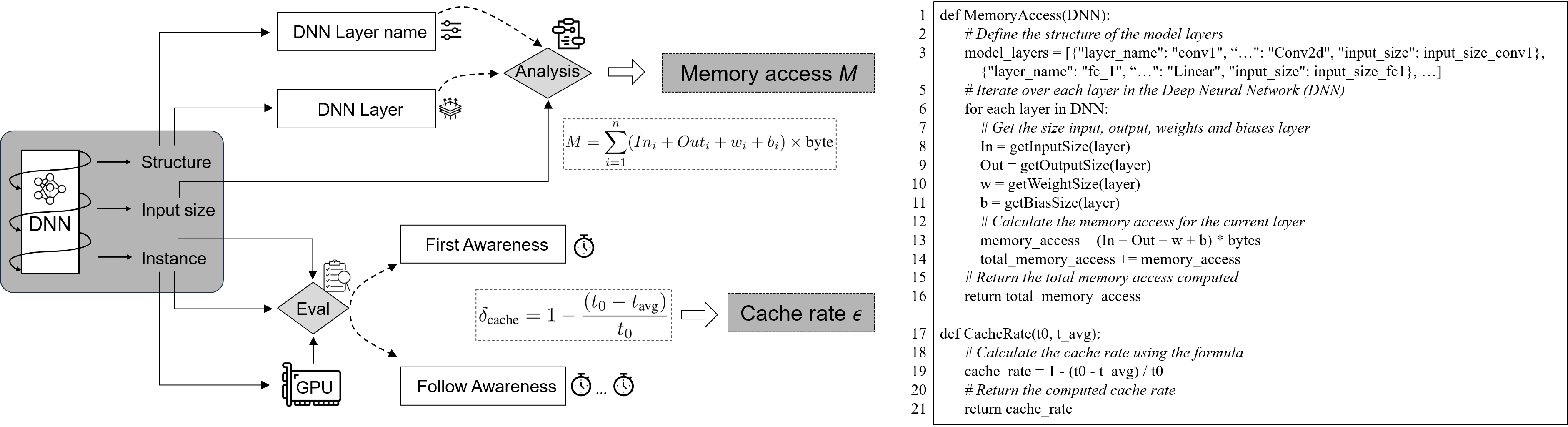}}
\caption{Left: a workflow of calculating memory access and cache rate. Right: pseudo codes in Python.}
\label{fig:cal-rate-M}
\end{figure*}

\subsection{Model Performance Profiler}
\label{Model Performance Profiler}

Evaluating DNNs requires a comprehensive analysis of performance metrics. These metrics can be divided into two categories: \textit{intrinsic performance} and \textit{predictive performance}.

\begin{itemize}
    \item \textit{Intrinsic performance} includes metrics related to the model's structure, such as computational complexity, parameter amount, and storage. These metrics are crucial for assessing the model's efficiency and are typically evaluated early in development, independent of the specific deployment device or training outcomes.
    \item \textit{Predictive performance} encompasses hardware-related metrics, focusing on latency and energy consumption. These metrics are vital for understanding how the model performs on mobile devices with varying resources.
\end{itemize}

A rapid and precise assessment of these performance metrics is essential for adapting DNNs to the dynamic computing environments of mobile devices. This early evaluation ensures that models are optimized for the constraints and capabilities of mobile platforms, leading to more efficient and effective deployment. This section describes how to evaluate these performance metrics in detail.

Evaluate the intrinsic performance metrics of deep learning models through the following aspects: \textit{computational cost ($C$)}, \textit{parameter amount ($P$)}, and \textit{storage ($S$)}. Computational Cost measures the total operations needed during training or inference, typically quantified by floating-point operations across convolutional, pooling, and fully connected layers. Parameter Amount assesses the cumulative learnable parameters, including weights and biases, layer by layer, indicating model complexity. Storage quantifies the required space to save a model by summing the sizes of all parameters, adjusted for their data types and configurations, often stored in formats such as .pth, .pt, or .onnx. These metrics collectively provide essential insights into a model’s resource demands, crucial for performance optimization in mobile and embedded systems.





\textit{Energy Consumption ($E$)}: this metric quantifies the electrical energy utilized by a model during its operation, integral for assessing predictive performance aspects like latency and energy consumption. It is crucial in mobile computing where both model architecture and device hardware impact energy demands. We calculate the energy consumption for a DNN using the following formula:

\begin{align} \label{eq:E}
    E = & \left( \left(M_{cpu} \times P_{cpu} \times \epsilon \right) \div f_{cpu} \right)  + \left( \left(M_{gpu} \times P_{gpu}\right) \div f_{gpu} \right) \nonumber \\
        & + M_{mem} \times P_{mem} \times (1-\epsilon) \div f_{mem},
\end{align}
where \( M_{cpu} \), \( M_{gpu} \), and \( M_{mem} \) represent the memory operations for the CPU, GPU, and general memory, respectively. \( \epsilon \) denotes cache hits. \( P_{cpu} \), \( P_{gpu} \), and \( P_{mem} \) are the power consumption rates for the CPU, GPU, and memory, respectively. Finally, \( f_{cpu} \), \( f_{gpu} \), and \( f_{mem} \) reflect the clock frequencies of the CPU, GPU, and memory, indicating the operations they can perform per second.



Our model considers a holistic view of energy requirements by accounting for computation and various memory operations, drawing on extensive empirical data from AdaEnlight~\cite{liu2023adaenlight} to estimate cache hit rates and energy coefficients. This comprehensive approach enables precise predictions and effective optimizations in the context of mobile computing. The \textit{memory access ($M$)} and \textit{cache hit rate (\(\epsilon\))} are carefully designed and calculated using the following method:

\begin{itemize}
    \item \textit{Memory accesses} is across various layers of a model, encompassing input, conv layers, pooling layers, and fc layers. It is quantified by the total number of memory accesses performed during the model's operation, which can be represented by the following formula:
    \begin{equation} \label{eq:M}
M = \sum_{i=1}^{n} (In_i + Out_i + w_i + b_i) \times \text{byte},
\end{equation}
where, \( In_i \), \( Out_i \), \( w_i \), and \( b_i \) represent the input, output, weights, and biases of each layer, respectively, each scaled by their memory consumption in bytes. 

    \item \textit{Cache hit rate} is often not directly visible for program operators. Therefore, we have devised a method to estimate it using access time to predict the cache rate. By analyzing access times, we can indirectly approximate the cache rate, providing a useful metric for optimizing performance in mobile computing environments.
\begin{equation} \label{eq:cache}
\delta_{\text{cache}} = 1 - (t_0 - t_{\text{avg}}) / {t_0}
\end{equation}
\end{itemize}

Specifically, As shown in Fig.~\ref{fig:cal-rate-M}, the memory access size for each layer can be calculated by multiplying the input, output, weights, and biases (if present) by their byte sizes. To compute the cache rate \(\epsilon\), we employ two methods: (i) For individual requests, the Single Request Hit Rate is determined by the ratio of cache hits to total requests, calculated as \textit{Hit Rate = (Number of Cache Hits) / (Total Number of Requests)}. (ii) For batch requests, the Batch Request Hit Rate is computed similarly, using \textit{Hit Rate = (Number of Successful Hits) / (Total Number of Requests)}. Additionally, we monitor access times at both the cache and storage layers to assess the cache rate, employing sampling and statistical methods to track and estimate the hit/miss status of each request. Initially, the cache rate is zero, but typically increases with subsequent operations, enhancing the precision of cache rate.

\textit{Latency ($L$):} we predict latency by integrating both theoretical estimates and actual device performance. This two-stage approach involves: \textit{theoretical latency} calculation is determined by analyzing the network parameters of different layer types and considering the device's processor architecture and clock frequency, i.e., the theoretical latency is calculated based on the amount of computations and the latency of each calculation; \textit{runtime latency} prediction evaluates the operating status of the device, collects resource utilization, and clock frequency under different load conditions, linking these parameters with actual runtime latency. By building models of theoretical and predicted latency, we can accurately forecast the inference latency of DNNs.



\subsection{Performance-Guided Search}
\label{Performance-Guided Search}



Deploying network architectures with multi-variant models and integrated compression techniques across diverse edge devices poses significant challenges. Traditional adaptive methods, such as those described in NAS~\cite{cai2018proxylessnas}, LegoDNN~\cite{han2021legodnn} and AdaSpring~\cite{liu2021adaspring}, often overlook the unique characteristics of each device, leading to inefficient and time-consuming adaptation processes for edge environments.

To address this problem, we introduce a approach that using a subspace search tailored to their specific needs. This device-adaptive phase includes a performance evaluation module (described in~\ref{Model Performance Profiler}), a resource availability awareness module(described in~\ref{Resource Availability Awareness}) and a model-guided search strategy, significantly reducing the search overhead and enabling rapid adaptation to the dynamic conditions of edge computing.







The primary goal of the model-guided search strategy is to effectively minimize the combined performance loss function \( J \), which is defined as:

\begin{equation} \label{eq:J}
J(m, d) = \alpha \cdot L(m, d) + \beta \cdot E(m, d),
\end{equation}
where \( \alpha \) and \( \beta \) are weights balancing latency and energy consumption. We aim to optimize under two constraints: latency \( L(m,d) \leq T \) and energy \( E(m,d) \leq E_b \). The objective is to find a model \( m' \) that minimizes \( J(m,d) \) while adhering to these constraints:

\begin{equation} \label{eq:m'}
\centering
\begin{aligned}
& m' = \text{arg min}_{m \in M} J(m, d), \\
& \text{s.t. } L(m, d) \leq T, \, E(m, d) \leq E_b
\end{aligned}
\end{equation}
which optimizes the model's performance by balancing latency and energy consumption, ensuring that it operates efficiently within device-specific constraints.




To identify the optimal model, we have constructed comprehensive tables for all network variants, encompassing both backbone networks and compression operators. Our \textit{Performance Index Table} catalogues each variant by the number of trainable parameters, storage, and accuracy. Additionally, the \textit{Predictive Performance Table} documents the latency and energy consumption for each configuration. These tables, derived from the intrinsic performance and accuracy metrics of DNNs post-training, ensure consistency across various deployment devices, providing a robust framework for model optimization.



Existing work has adopted specific data structures to accelerate searching.  AdaSpring~\cite{liu2021adaspring} uses common prefixes and organizes the search space into a navigable structure, enhancing the speed of sub-block searches. Similarly, AdaptiveNet~\cite{wen2023adaptivenet} uses a candidate table that classifies sub-models by latency intervals, streamlining model selection. These methods simplify the optimization process by using comprehensive index tables that pre-store DNN performance data. 

For our performance-guided search, we leverage pre-stored model performance data to swiftly retrieve and compare model variants, enabling the efficient identification of the optimal model based on performance, latency, and energy criteria. Unlike traditional methods such as AdaSpring and AdaptiveNet, which are less suitable due to our use of lightweight compression operators and a smaller search space, we employ a B+ tree structure to manage data across two tables, optimizing for rapid access and effective range searches. 

Our method is particularly advantageous in handling large volumes of network variant data in real-time. The B+ tree's leaf-node chaining facilitates efficient range queries, ensuring quick multi-dimensional model performance comparisons, while its stability supports dynamic updates with new network variants. Additionally, by integrating intrinsic and predictive performance tables, the B+ tree structure aids in pinpointing models that meet specific criteria, enhancing our capability to efficiently identify and retrieve optimal network variants that align with our defined performance constraints.

\section{EVALUATION}
\label{EVALUATION}


We aim to address the following research questions through experimentation:
$Q_1:$ Can AdaScale achieve models with better trade-offs between latency and accuracy?
$Q_2:$ Is AdaScale capable of adapting to the available resources of IoT devices?
$Q_3:$ Does AdaScale exhibit better performance in dynamic scenarios? We will compare AdaScale with various methods reported in the latest literature to evaluate its effectiveness.

\subsection{Experiment Setup}
\label{Experiment-Setup}

We first present the settings for our evaluation.

\textbf{System Implementation.} We implemented our method using Python and C++. The training and deployment parts of the deep learning model are implemented using Python and PyTorch, while the device-side computation resource sensing and real-time monitoring are handled using C++. We have implemented the online component of AdaScale on mobile devices to dynamically adjust the DNN configuration to match the device's computational capabilities. The self-evolving networks generated by AdaScale's offline component (i.e., the backbone network and multiple compression operators) are then loaded into the target platform.

\begin{table}[t]
\caption{Summary of datasets for evaluating AdaScale}
\begin{center}
\begin{tabular}{|l|l|l|ll}
\cline{1-3}
\textbf{No.} & \textbf{Dataset} & \textbf{Description}                     &  &  \\ \cline{1-3}
D1           & Cifar10~\cite{krizhevsky2009learning}          & 60000 32×32 color images of 10 classes   &  &  \\ \cline{1-3}
D2           & Cifar100~\cite{krizhevsky2009learning100}          & 60000 32×32 color images of 100 classes  &  &  \\ \cline{1-3}
D3           & Tiny ImageNet~\cite{le2015tiny}    & 100000 64×64 color images of 200 classes &  &  \\ \cline{1-3}
\end{tabular}
\label{tab:dataset}
\end{center}
\end{table}

\textbf{Evaluation Applications/Datasets.} We utilize three commonly used mobile datasets to assess the performance of AdaScale, as shown in Table~\ref{tab:dataset}, which represents tasks of different difficulty levels. Specifically, we test AdaScale on mobile image classification tasks (D1: Cifar10~\cite{krizhevsky2009learning}, D2: Cifar100~\cite{krizhevsky2009learning100}, D3: Tiny ImageNet~\cite{le2015tiny}).

\textbf{Mobile Platforms with Dynamic Context Settings.} We evaluated AdaScale on three commonly used mobile and embedded platforms: an edge server \ie NVIDIA GeForce RTX 3080 with 10 GB GPU memory, an embedded development board \ie Raspberry Pi 4B, and a mobile robotic platform \ie 
 Turtlebot (Jeton Nx) with a mobile development board. These platforms are equipped with various processors, storage capacities, and battery capacity.

\textbf{Comparison Baselines.} We implemented and compared four state-of-the-art model scaling generation techniques from the literature. This includes LegoDNN~\cite{han2021legodnn} and Adaptivenet~\cite{wen2023adaptivenet}, which utilize block generation supernetwork techniques for network compression, as well as AdaDeep~\cite{liu2020adadeep} and AdaSpring~\cite{liu2021adaspring}, which employ selection and combination of model compression techniques to generate specialized networks that balance latency and accuracy.  In addition, we also use Neural Network Search (NAS)~\cite{zoph2016neural} as a basic baseline. 

\begin{itemize}
    \item LegoDNN~\cite{han2021legodnn} is a pruning-based block granularity model scaling technique.
    \item AdaptiveNet~\cite{wen2023adaptivenet} is a post-deployment model scaling technique that adjusts the model architecture after being deployed to the target environment.
    \item AdaDeep~\cite{liu2020adadeep} automatically selects and combines compression techniques to generate specialized DNNs that balance accuracy and resource constraints.
    \item AdaSpring~\cite{liu2021adaspring} is a latency budget model scaling technique based on various types of compression operators.
    \item NAS~\cite{zoph2016neural} improves model performance by automatically searching for the optimal network architecture.
\end{itemize}

\begin{figure*}[t]
    \centering
    \begin{subfigure}[b]{0.32\textwidth}
        \includegraphics[width=\textwidth]{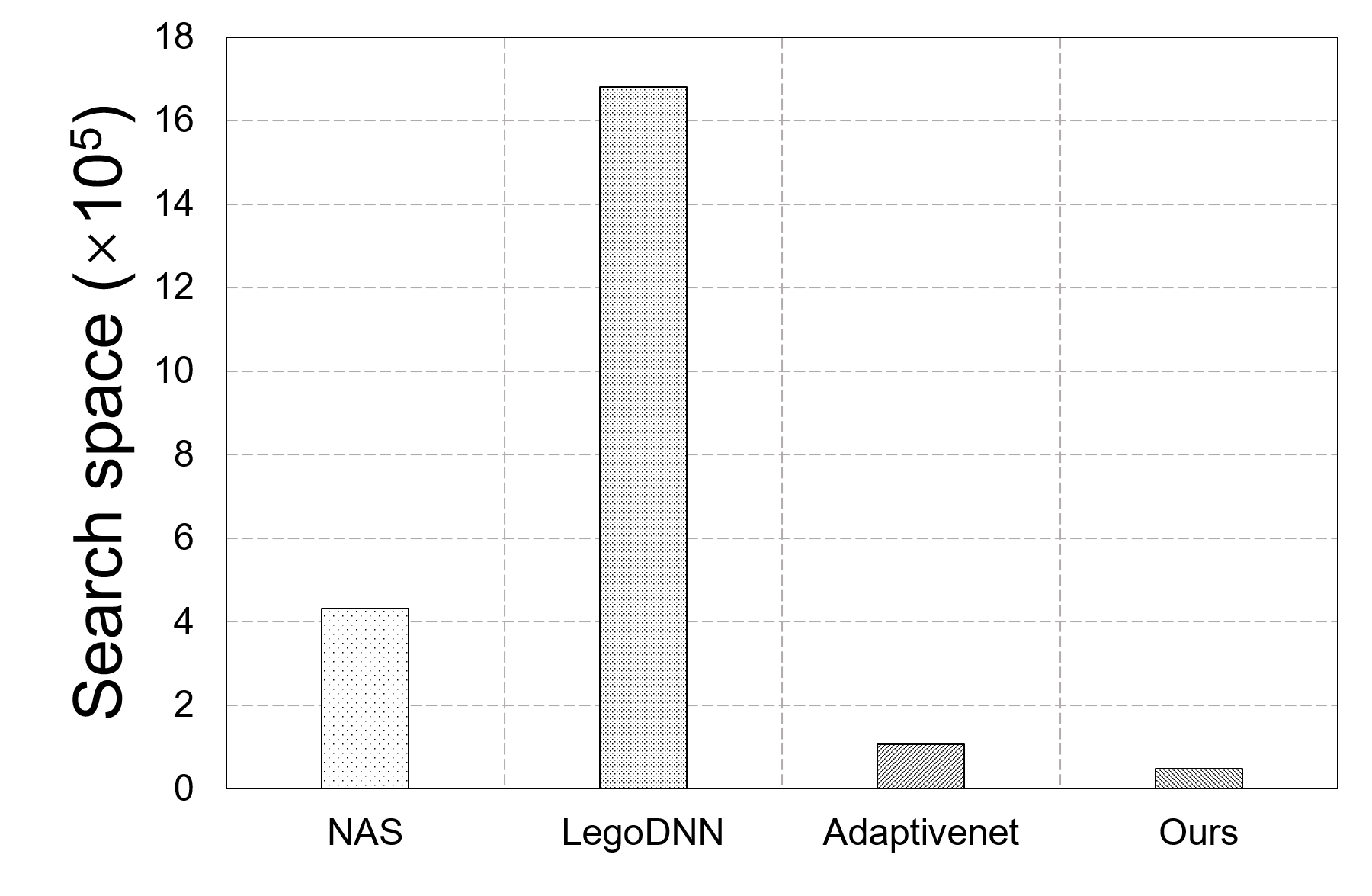}
        \caption{Search space}
        \label{fig:fig_serach_space}
    \end{subfigure}
    \hfill
    \begin{subfigure}[b]{0.32\textwidth}
        \includegraphics[width=\textwidth]{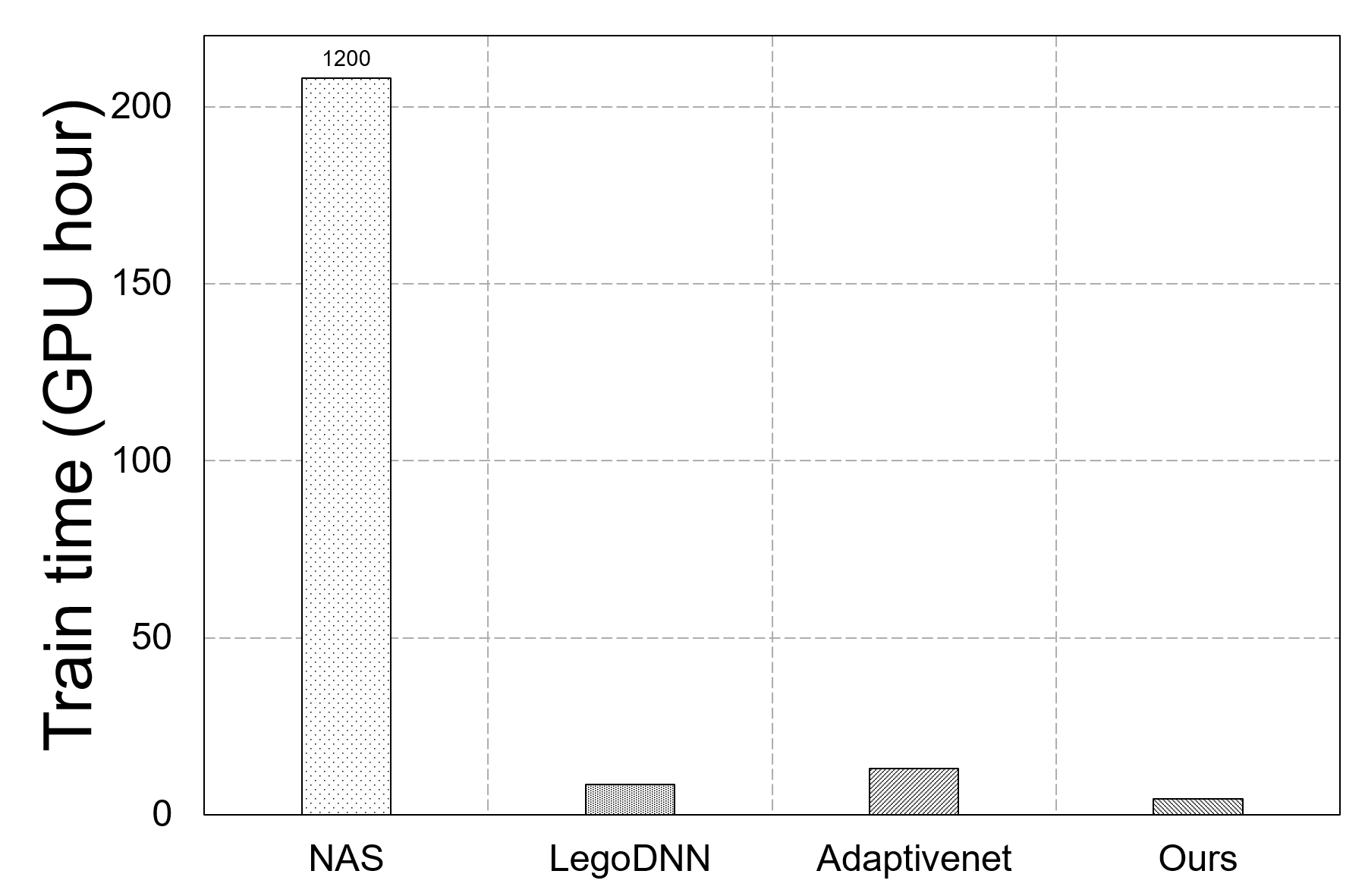}
        \caption{Training time}
        \label{fig:fig_train_time}
    \end{subfigure}
    \hfill
    \begin{subfigure}[b]{0.33\textwidth}
        \includegraphics[width=\textwidth]{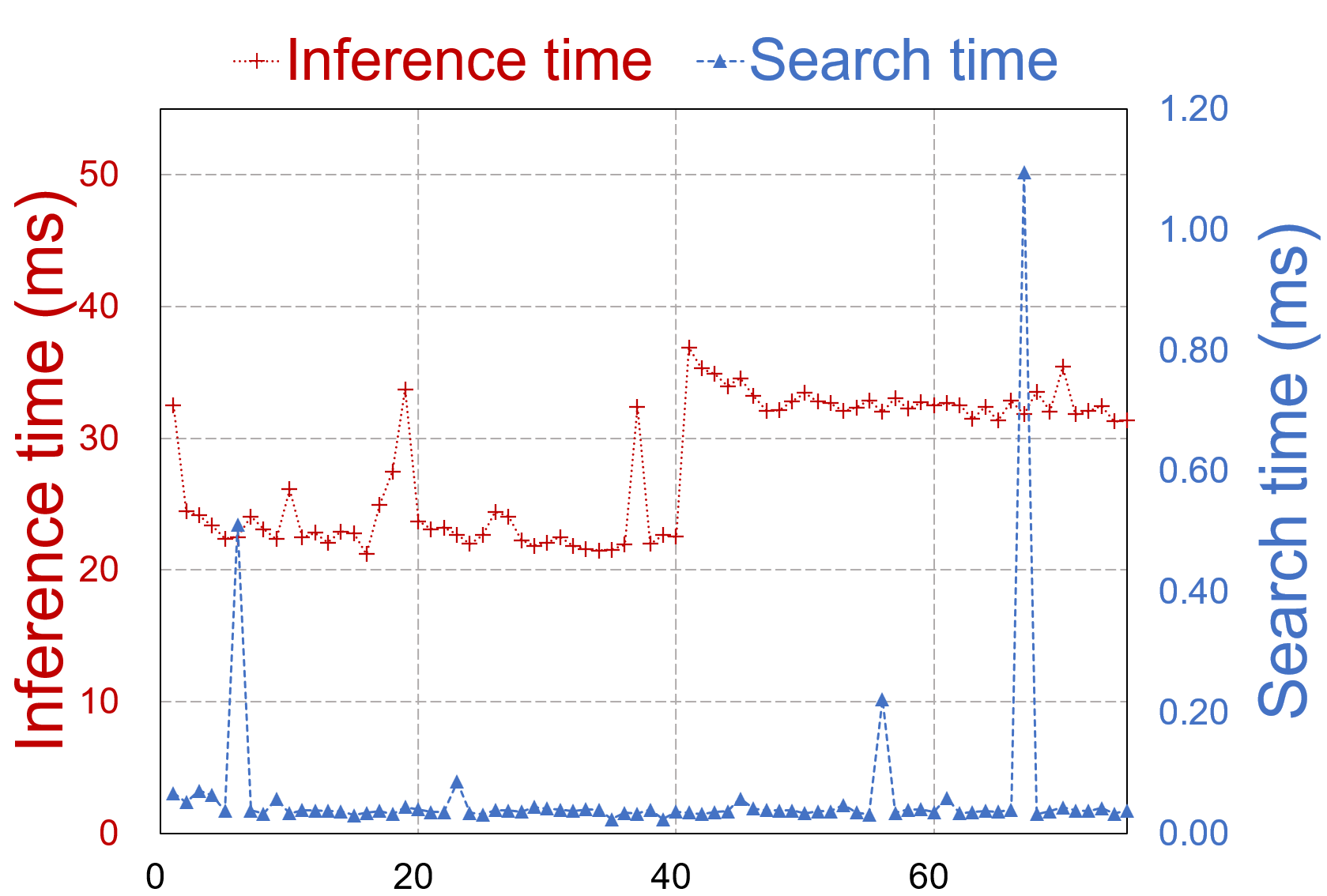}
        \caption{Latency}
        \label{fig:Search_time_Inference_time}
    \end{subfigure}
    \caption{AdaScale improves DNN efficiency by optimizing search space, training time, and search overhead.}
\end{figure*}

To better evaluate the effectiveness and advancement of our method, we compared AdaScale with established lightweight networks designed for mobile devices, including MobileNet~\cite{howard2017mobilenets}, Shufflenet~\cite{zhang2018shufflenet}, Squeezenet~\cite{iandola2016squeezenet}, and ResNet18~\cite{he2016deep}. Additionally, we also contrasted our results with the promising recent Transformer~\cite{vaswani2017attention} models to gauge performance across different architectural paradigms.

The lightweight single networks baseline:
\begin{itemize}
    \item MobileNet~\cite{howard2017mobilenets} utilizes depthwise separable convolutions for mobile and embedded vision applications. 
    \item Shufflenet~\cite{zhang2018shufflenet} enhances operational efficiency on mobile devices by optimizing the network architecture through a channel shuffle strategy. 
    \item Squeezenet~\cite{iandola2016squeezenet} significantly reduces the number of model parameters with its Fire modules. 
    \item ResNet18~\cite{he2016deep} employs residual connections to improve the performance of deep networks.
    \item Transformer~\cite{vaswani2017attention} employs a self-attention mechanism to process image patches, enabling it to effectively capture both global and local feature information.
\end{itemize}

We will conduct a performance comparison of AdaScale from four aspects: comparing with dynamic adaptive networks assessing search space, training time, and search overhead; comparing with lightweight single-network models focusing on accuracy, training memory overhead, training time, storage, inference latency, and energy consumption; evaluating AdaScale itself concentrating on the number of branches, storage overhead, inference latency, and resource adaptability; and testing AdaScale across different deployment scenarios. This structured approach allows us to comprehensively evaluate AdaScale from multiple perspectives.

\subsection{Performance Comparison of  Dynamic Adaptive Networks}
\label{Performance Comparison of  Dynamic Adaptive Networks}

We evaluate the quality of models generated by our method compared to generative and dynamic adaptive baseline models. Specifically, we assess the search space, training time, and deployment search time on mobile devices for three different generative network approaches: NAS, AdaptiveNet, and LegoDNN. To ensure generalizability and consistency with other studies, we use ResNet as our backbone model, as it is commonly employed in the experimental setups of AdaptiveNet, LegoDNN, and AdaSpring.

\begin{table}[t]
\centering
\caption{Performance comparison of dynamic adaptive networks}
\begin{tabular}{|c|c|c|c|}
\hline
\textbf{Methods} & Overhead  & Train Time/GPU hour & Search Space       \\ \hline
NAS~\cite{zoph2016neural}              & 196 hours & 1200                & 4.32 × 10$^5$           \\ \hline
Adaptivenet~\cite{wen2023adaptivenet}      & 117.6 s   & 13                  & 1.06 × 10$^5$ \\ \hline
LegoDNN~\cite{han2021legodnn}          & $>$ 1 s     & 8.6                 & 1.68 × 10$^6$           \\ \hline
AdaSpring~\cite{liu2021adaspring}        & 3.8 ms    & -                   & -                  \\ \hline
AdaDeep~\cite{liu2021adaspring}          & 18 hours  & -                   & -                  \\ \hline
Ours             & \textbf{0.043 ms}  & \textbf{4.575}               & \textbf{0.48 × 10$^3$}           \\ \hline
\end{tabular}
\label{tab:Comparison_Dynamic}
\end{table}

\textbf{Performance.} \textit{(i)} As illustrated in Figure~\ref{fig:fig_serach_space}, using lightweight network modules as our compression operators significantly reduces our search space compared to state-of-the-art network generation methods~\cite{han2021legodnn,wen2023adaptivenet,zoph2016neural}. AdaScale's space is reduced to 0.48×10$^{3}$, which is one percent of that of AdaptiveNet and one thousandth of LegoDNN, yet achieves comparable results. This reduction minimizes the resources required to generate networks.
\textit{(ii)} As shown in Figure~\ref{fig:fig_train_time}, the smaller search space results in the shortest training time for our network. Despite the higher computational power of the servers used by NAS, LegoDNN, and AdaptiveNet, AdaScale demonstrates greater potential.
\textit{(iii)} In our deployment of AdaScale on the TurtleBot robot, as shown in Table~\ref{tab:Comparison_Dynamic} and Figure~\ref{fig:Search_time_Inference_time}, the runtime search overhead was only 0.043ms, approximately one percent of the average inference latency. This efficiency is due to our model performance evaluation and device resource sensing modules. The performance table uses a B+ tree data structure, ensuring each query is processed in $O(\log_b n)$ time, where $b$ is the branching factor and $n$ is the total number of elements (0.48×10$^{3}$). This approach significantly reduces search and evaluation costs.

\textbf{Summary:} AdaScale dramatically reduces the search space from one thousandth to one percent of more advanced methods. This substantial reduction not only decreases the resources needed for network generation but also significantly shortens training time. In practical applications, AdaScale shows exceptionally low runtime search overhead on the TurtleBot robot, significantly enhancing both efficiency and performance.

\begin{table*}[!t]
\begin{center}
\scriptsize
\caption{Performance of different lightweight models on Cifar10}

\begin{tabular}{|cc|c|c|c|c|c|c|c|}
\hline
\multicolumn{2}{|c|}{\textbf{Methods}}               & \textbf{Top1 Acc(\%)} & \textbf{Training Memory (MB)} & \textbf{Training Time (hour)} & \textbf{Storage (MB)} & \textbf{Parameter(M)} & \textbf{Latency(ms)} & \textbf{Energy(mJ)} \\ \hline
\multicolumn{2}{|c|}{Shufflenet}                     & 89.73                & 1468                          & 1.17                          & 3.74                  & 1.01                  & 29.70                & 5185.79             \\ \hline
\multicolumn{2}{|c|}{Squeezenet}                     & 87.86                & 1212                          & 0.68                          & 2.88                  & 0.78                  & 16.80                & 4399.88             \\ \hline
\multicolumn{2}{|c|}{Resnet18}                       & 88.61                & 1314                          & 1.02                          & 42.70                 & 11.22                 & 16.60                & 44636.81            \\ \hline
\multicolumn{2}{|c|}{Mobilenetv3large}               & 78.85                & 880                           & 1.33                          & 15.15                 & 4.03                  & 34.00                & 412.08              \\ \hline
\multicolumn{2}{|c|}{Mobilenetv3small}               & 74.18                & 752                           & 1.17                          & 6.58                  & 1.80                  & 28.00                & 120.61              \\ \hline
\multicolumn{2}{|c|}{Transformer}                    & 90.01                 & 8442                          & 7.77                          & 71.89                 & 12.44                 & 21.84                & 2602781.36          \\ \hline
\multicolumn{1}{|c|}{\multirow{4}{*}{Ours}} & Stage1 & 85.99                & 978                           & 0.42                          & \textbf{1.04}         & \textbf{0.26}         & \textbf{9.10}        & 1285.17             \\ \cline{2-9} 
\multicolumn{1}{|c|}{}                      & Stage2 & 87.77                & 806                           & \textbf{0.32}                 & 5.25                  & 1.36                  & 13.90                & 2584.63             \\ \cline{2-9} 
\multicolumn{1}{|c|}{}                      & Stage3 & \textbf{91.17}       & 494                           & 0.48                          & 6.50                  & 1.68                  & 19.20                & 2692.62             \\ \cline{2-9} 
\multicolumn{1}{|c|}{}                      & Stage4 & 90.82                & \textbf{486}                  & 0.37                          & 9.65                  & 2.50                  & 23.70                & 2758.59             \\ \hline
\end{tabular}
\label{tab:cifar10withmodels}
\end{center}
\end{table*}

\begin{table*}[h]
\scriptsize
\caption{Performance of different lightweight models on Cifar100}
\begin{center}
\begin{tabular}{|cc|c|c|c|c|c|c|c|}
\hline
\multicolumn{2}{|c|}{\textbf{Methods}}               & \textbf{Top1 Acc(\%)} & \textbf{Training Memory (MB)} & \textbf{Training Time (hour)} & \textbf{Storage (MB)} & \textbf{Parameter(M)} & \textbf{Latency(ms)} & \textbf{Energy(mJ)} \\ \hline
\multicolumn{2}{|c|}{Shufflenet}                     & 70.99                & 1534                          & 1.16                          & 4.07                  & 1.01                  & 29.30                & 5185.79             \\ \hline
\multicolumn{2}{|c|}{Squeezenet}                     & 69.55                & 1218                          & 0.55                          & 3.06                  & 0.78                  & 17.10                & 4399.88             \\ \hline
\multicolumn{2}{|c|}{Resnet18}                       & 75.31                & 1316                          & 0.91                          & 42.88                 & 11.22                 & 16.50                & 44636.81            \\ \hline
\multicolumn{2}{|c|}{Mobilenetv3large}               & 51.59                & 880                           & 1.33                          & 15.59                 & 4.03                  & 33.80                & 412.08              \\ \hline
\multicolumn{2}{|c|}{Mobilenetv3small}               & 46.02                & 752                           & 1.16                          & 7.02                  & 1.80                  & 27.80                & 120.61              \\ \hline
\multicolumn{2}{|c|}{Transformer}                    & 66.32                     & 8454                          & 7.30                              & 72.29                 & 12.44                 & 21.41                & 2602781.36          \\ \hline
\multicolumn{1}{|c|}{\multirow{4}{*}{Ours}} & Stage1 & 66.99                & 978                           & 0.51                          & \textbf{2.44}         & \textbf{0.26}         & \textbf{9.10}        & 1285.17             \\ \cline{2-9} 
\multicolumn{1}{|c|}{}                      & Stage2 & 69.83                & 810                           & \textbf{0.42}                 & 5.96                  & 1.36                  & 13.80                & 2584.63             \\ \cline{2-9} 
\multicolumn{1}{|c|}{}                      & Stage3 & \textbf{72.12}       & 496                           & 0.51                          & 6.59                  & 1.68                  & 19.20                & 2692.62             \\ \cline{2-9} 
\multicolumn{1}{|c|}{}                      & Stage4 & 71.95                & \textbf{488}                  & 0.49                          & 9.69                  & 2.50                  & 23.70                & 2758.59             \\ \hline
\end{tabular}
\label{tab:cifar100withmodels}
\end{center}
\end{table*}

\subsection{Comprehensive Performance Comparison}
\label{Performance Comparison of Lightweight Single-Network}

We evaluate our approach (with a 4-stage network) against lightweight single-networks, selecting Shufflenet, Squeezenet, ResNet18, MobileNetv3 (small and large) and Transformer for comparison. Our baseline training configuration is as follows: batch size of 128, 16 workers, using SGD as the optimizer with a momentum of 0.9 and weight decay of 5e-4. MILESTONES are set at epochs 40, 80, 100, and 120, with an initial learning rate of 0.1. We test these methods and our approach on two datasets, Cifar10 and Cifar100, evaluating accuracy, training memory, training time, storage, parameters, deployment runtime latency, and energy consumption on Jetson Nx.

\textbf{Performance.} \textit{(i)} As shown in Table~\ref{tab:cifar10withmodels}, our four-stage segmented training method significantly enhances model performance on the CIFAR-10 dataset. AdaScale's accuracy progressively improves, starting at 85.99\% and peaking at 91.17\% in the third stage, before stabilizing at 90.82\% in the fourth. This progression underscores its effective adaptation and learning capabilities. Notably, AdaScale surpasses the peak accuracy of the Transformer model, which is 90.0\%, while requiring only one-tenth of the Transformer’s training memory and time. Additionally, AdaScale outperforms both versions of Squeezenet and Mobilenetv3 in the later stages. In terms of training resource consumption, AdaScale consistently requires less memory across all stages compared to Shufflenet and ResNet18, reaching a low of 978MB. It also demonstrates faster training speeds, decreasing from 0.42 hours in the first stage to 0.37 hours in the fourth—significantly shorter than Mobilenetv3large’s 1.33 hours. Our model's storage needs modestly increase from 1.04MB in the first stage to 9.65MB in the fourth, which is significantly lower than ResNet18’s 42.70MB. The model's parameters increase from 0.26 million in the first stage to 2.50 million by the fourth stage. This reflects a transition from simpler to more complex structures that enhance performance and accuracy, making it suitable for resource-limited environments. Additionally, AdaScale maintains low latency across various mobile devices, especially in the first stage, in contrast to other baselines, which exhibit unstable inference latencies. AdaScale shows remarkable efficiency, recording the lowest energy consumption at 1285.17 millijoules in the first stage. As the number of parameters grows in later stages, the increase in energy consumption is kept moderate, effectively balancing performance enhancements with energy usage.

\textit{(ii)} As shown in Table~\ref{tab:cifar100withmodels}, a comparison on the CIFAR-100 dataset illustrates the effectiveness of our segmented training method for elastic networks. AdaScale achieved its highest accuracy of 72.12\% in the third stage, highlighting the effectiveness of our training strategy. ResNet18 leads with an accuracy of 75.31\%, but AdaScale surpasses both Shufflenet and Squeezenet and approaches ResNet18's performance. Regarding training resource consumption, AdaScale's memory requirements significantly decreased from 978MB in the first stage to 488MB in the fourth, illustrating a substantial reduction as training progressed. Training time also improved, dropping from 0.51 hours in the first stage to 0.49 hours in the fourth, enhancing training efficiency. The storage needs of AdaScale increased gradually from 2.44MB in the first stage to 9.69MB in the fourth but remained considerably lower than ResNet18's 42.88MB. Despite this growth in storage requirements, the increase was efficiently managed. However, latency and energy consumption rose from 9.10ms and 1285.17mJ in the first stage to 23.70ms and 2758.59mJ in the fourth, reflecting the model’s increased complexity. This increase in performance costs is deemed acceptable due to the significant improvements in accuracy.

\textbf{Summary:} AdaScale significantly improves DNN performance, surpassing baseline accuracy during its phased training process. It also requires less memory and training time than baseline. Storage needs and parameter count increase through each phase but are substantially lower than ResNet18's, while maintaining acceptable energy efficiency. This balance makes AdaScale ideal for resource-constrained environments.

\begin{figure*}[t]
    \centering
    \begin{subfigure}[b]{0.3\textwidth}
        \centering
        \includegraphics[width=\textwidth]{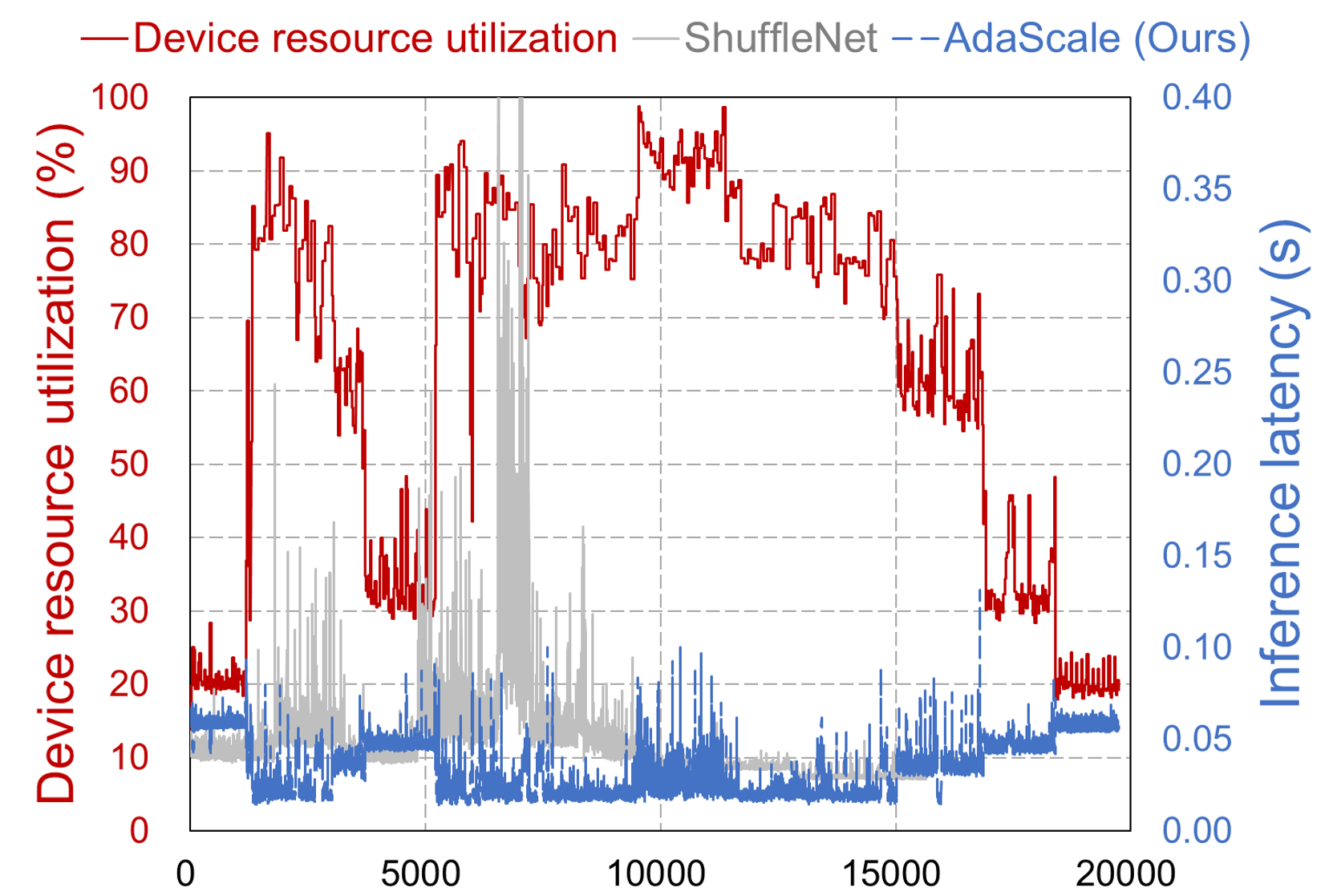}
        \caption{ShuffleNet on Jetson Nx}
        \label{fig:ShuffleNet}
    \end{subfigure}
    \hfill
    \begin{subfigure}[b]{0.3\textwidth}
        \centering
        \includegraphics[width=\textwidth]{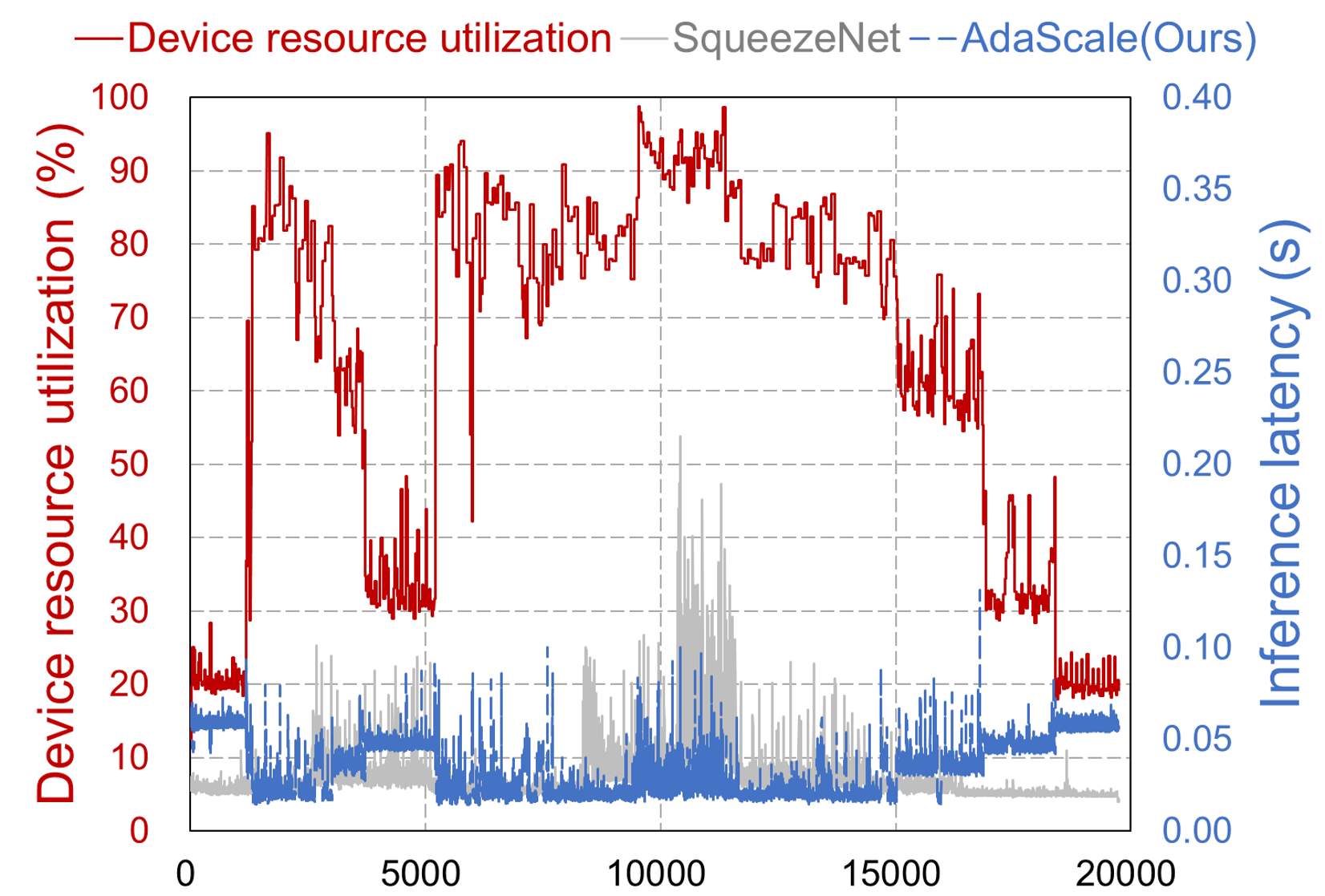}
        \caption{SqueezeNet on Jetson Nx}
        \label{fig:SqueezeNet}
    \end{subfigure}
    \hfill
    \begin{subfigure}[b]{0.3\textwidth}
        \centering
        \includegraphics[width=\textwidth]{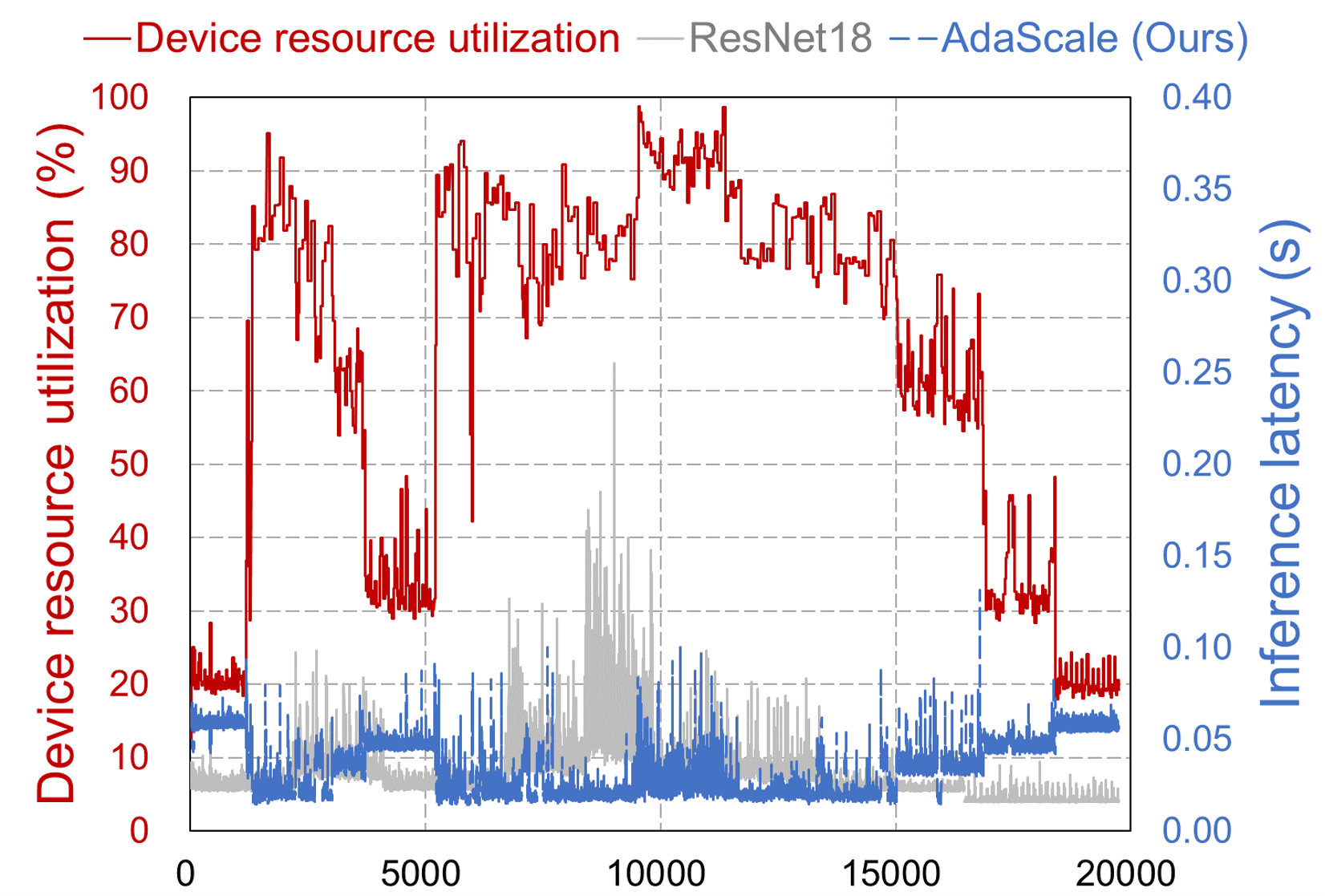}
        \caption{ResNet18 on Jetson Nx}
        \label{fig:ResNet18}
    \end{subfigure}
    
    \begin{subfigure}[b]{0.3\textwidth}
        \centering
        \includegraphics[width=\textwidth]{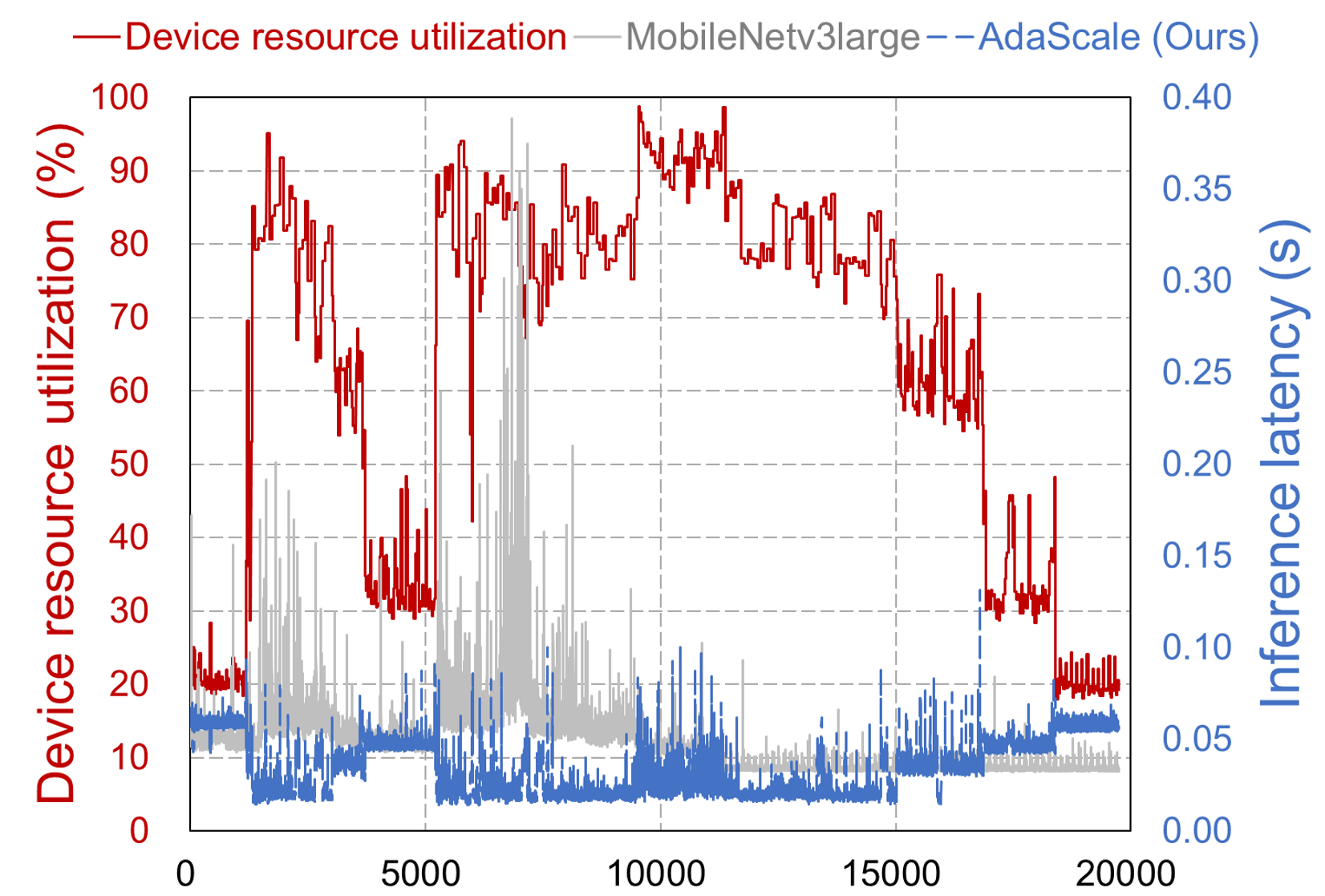}
        \caption{MobileNet V3 large on Jetson Nx}
        \label{fig:MobileNet V3 large}
    \end{subfigure}
    \hfill
    \begin{subfigure}[b]{0.3\textwidth}
        \centering
        \includegraphics[width=\textwidth]{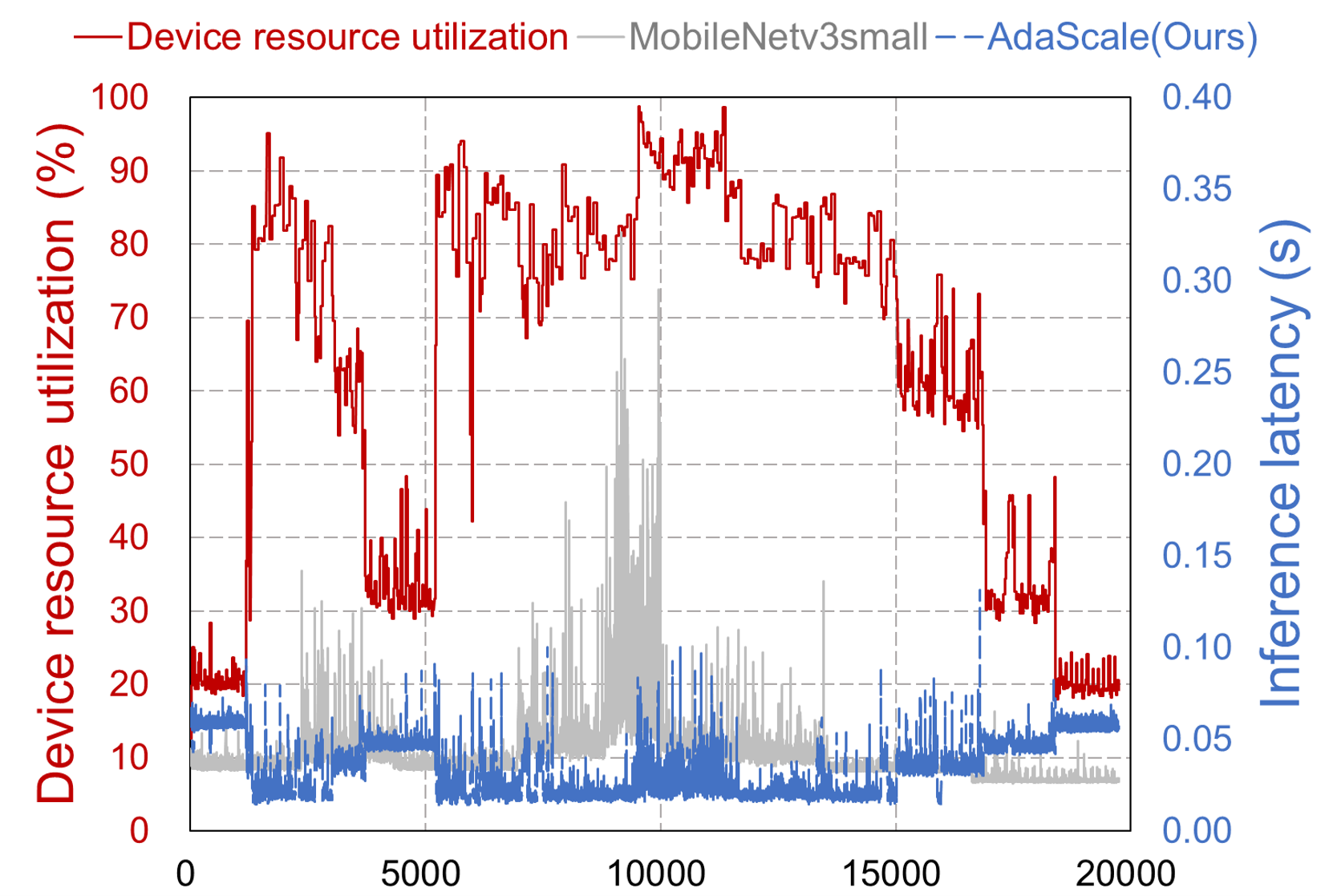}
        \caption{MobileNet V3 small on Jetson Nx}
        \label{fig:MobileNet V3 large}
    \end{subfigure}
    \hfill
    \begin{subfigure}[b]{0.3\textwidth}
        \centering
        \includegraphics[width=\textwidth]{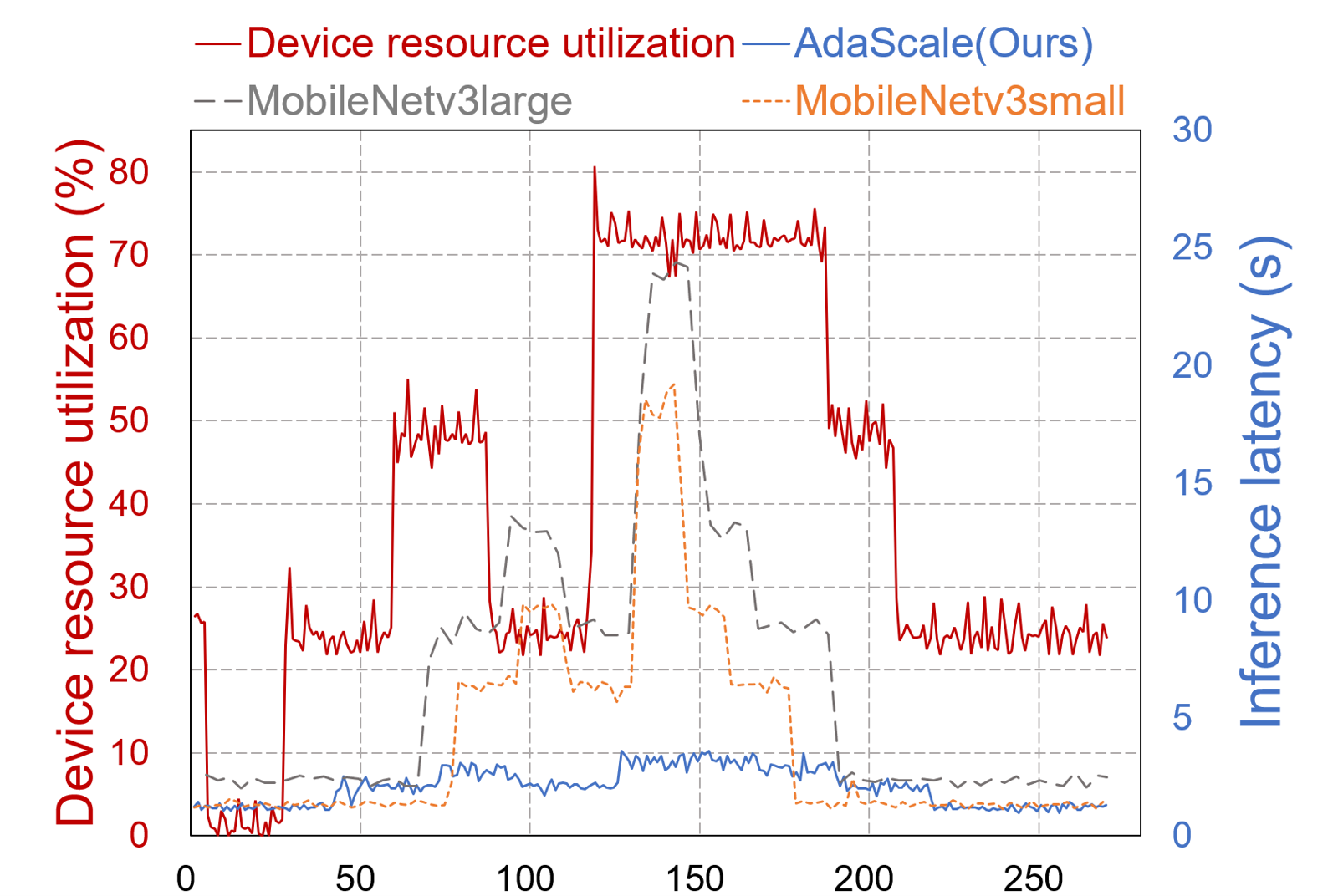}
        \caption{MobileNet V3 on Raspberry Pi 4B}
        \label{fig:MobileNet V3 pi}
    \end{subfigure}

    \begin{subfigure}[b]{0.3\textwidth}
        \centering
        \includegraphics[width=\textwidth]{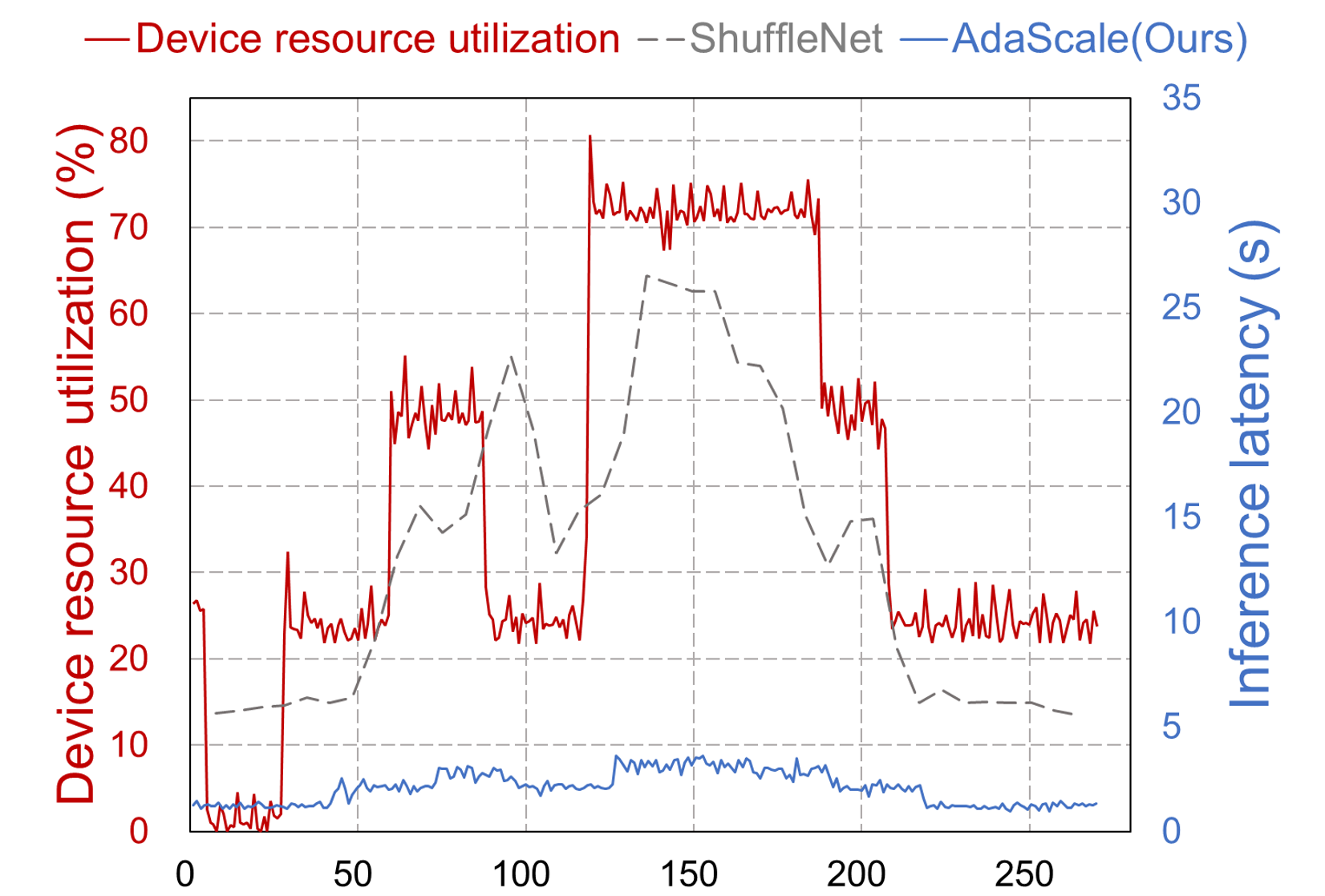}
        \caption{ShuffleNet on Raspberry Pi 4B}
        \label{fig:ShuffleNet pi}
    \end{subfigure}
    \hfill
    \begin{subfigure}[b]{0.3\textwidth}
        \centering
        \includegraphics[width=\textwidth]{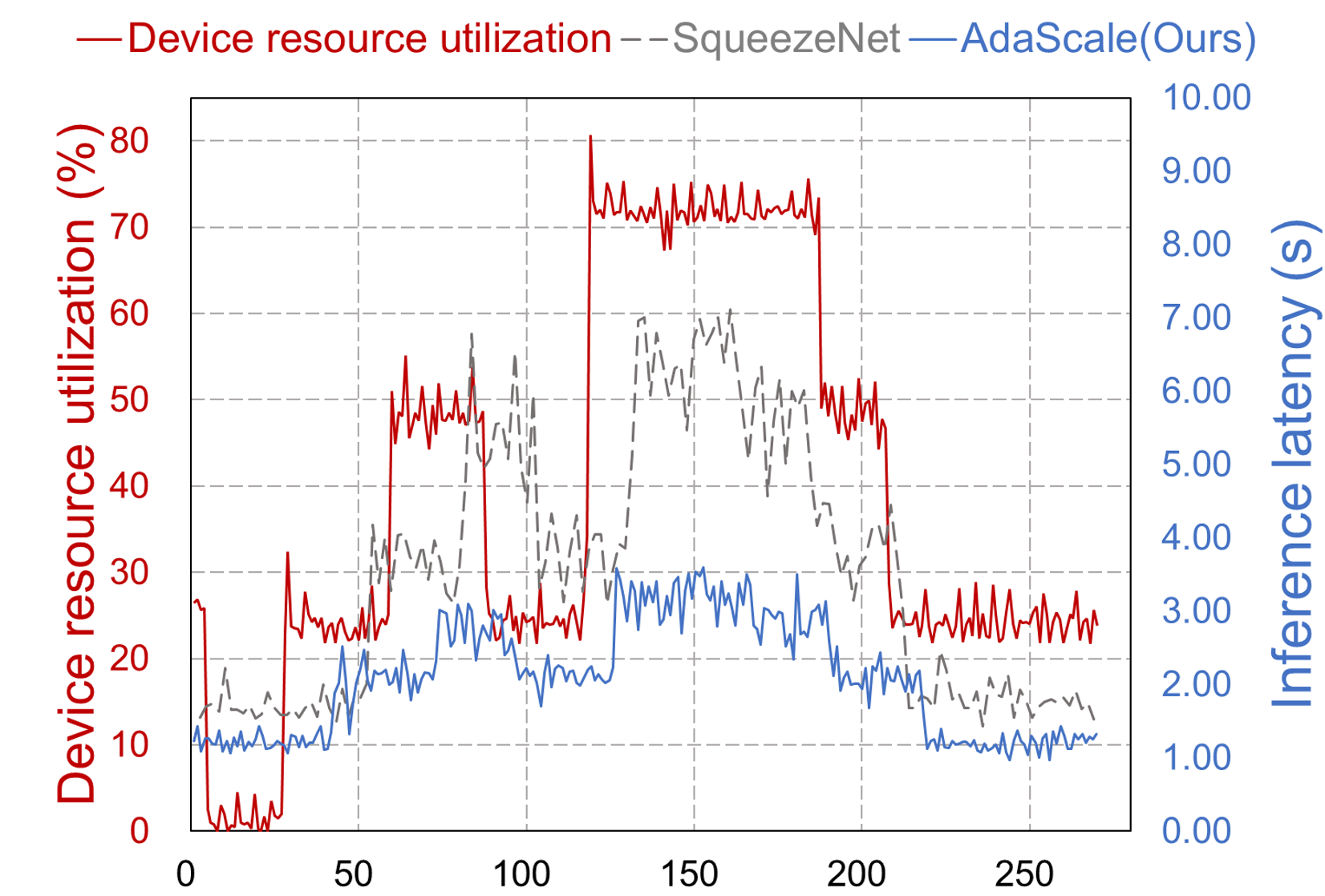}
        \caption{SqueezeNet on Raspberry Pi 4B}
        \label{fig:SqueezeNet pi}
    \end{subfigure}
    \hfill
    \begin{subfigure}[b]{0.3\textwidth}
        \centering
        \includegraphics[width=\textwidth]{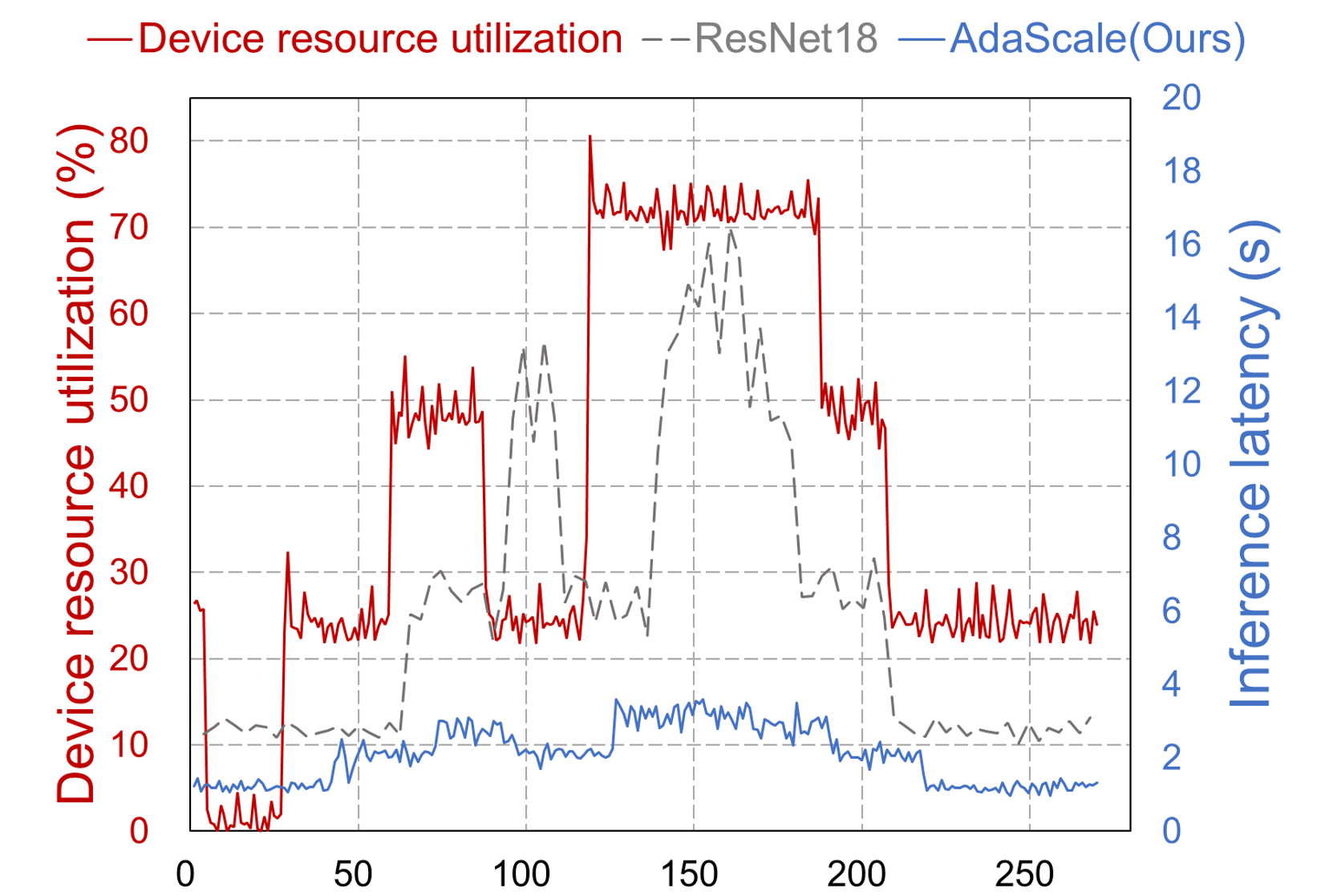}
        \caption{ResNet18 on Raspberry Pi 4B}
        \label{fig:ResNet18 pi}
    \end{subfigure}

    \caption{Inference latency of different DNNs under dynamic context on mobile devices}
    \label{fig:dynamic context}
\end{figure*}

\begin{figure*}[!h]
    \centering
    \begin{subfigure}[b]{0.3\textwidth}
        \centering
        \includegraphics[width=\textwidth]{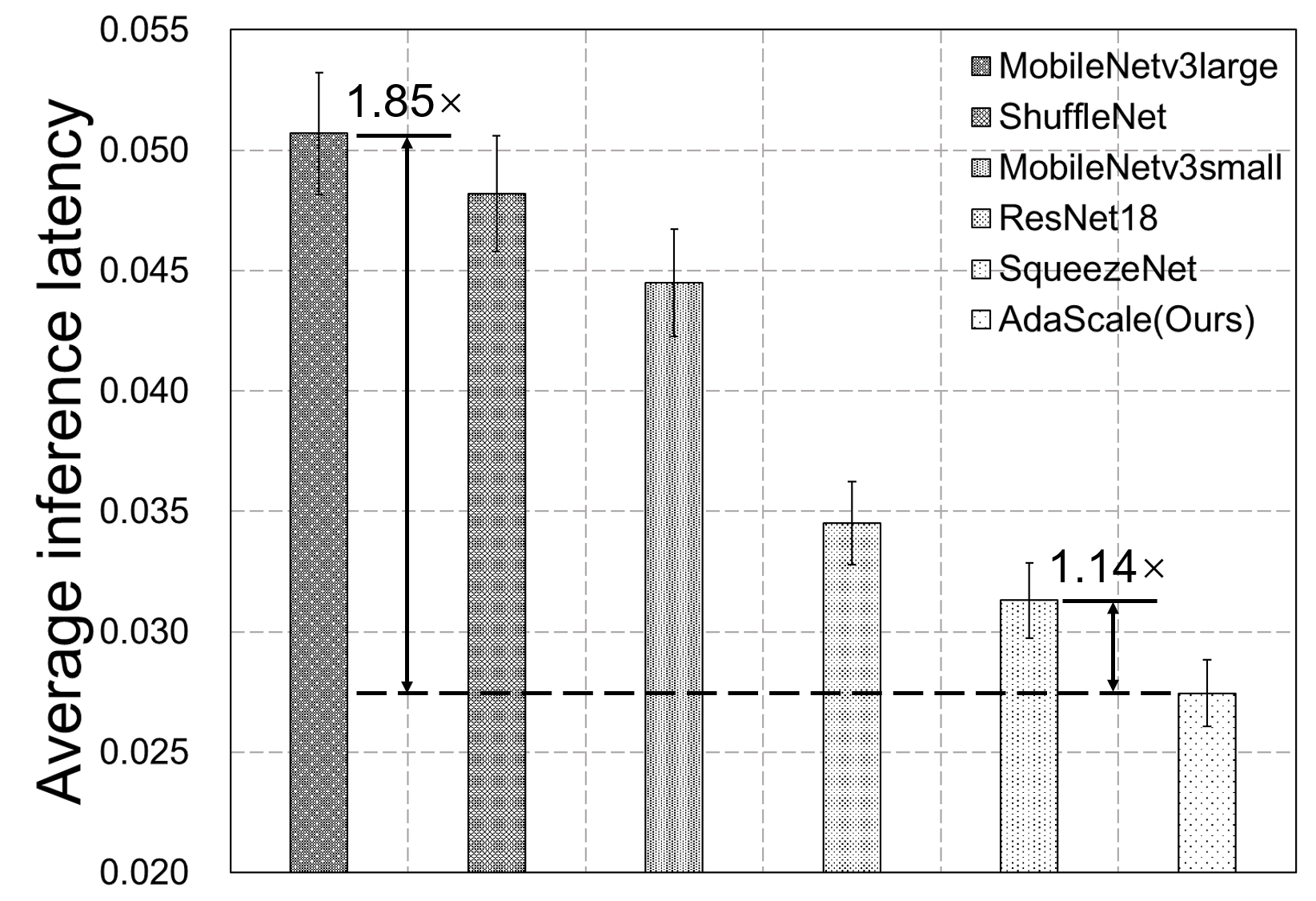}
        \caption{Average latency on Jetson Nx}
        \label{fig:jetson-cifar10-avg-latency}
    \end{subfigure}
    \hfill
    \begin{subfigure}[b]{0.3\textwidth}
        \centering
        \includegraphics[width=\textwidth]{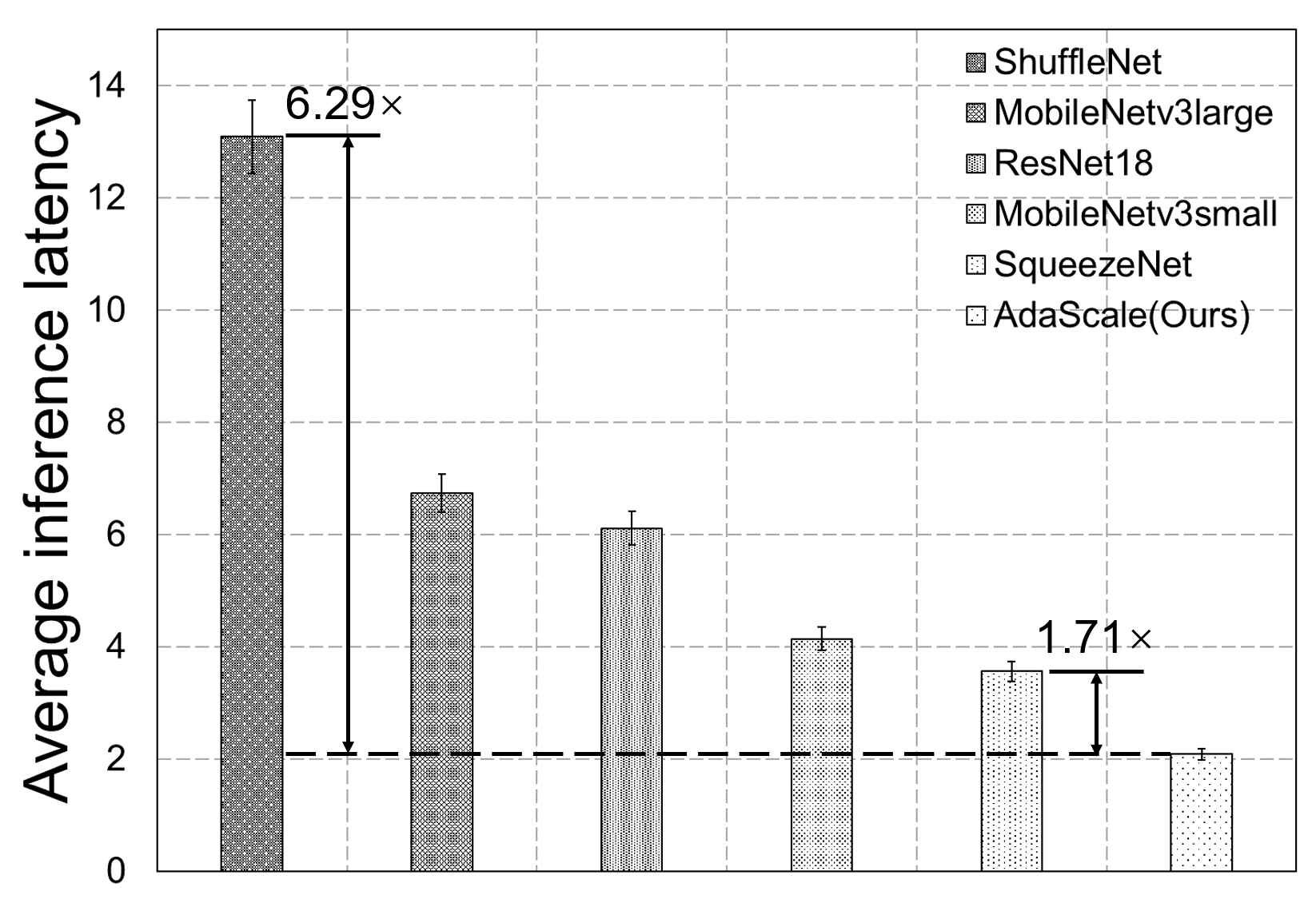}
        \caption{Average latency on Raspberry Pi 4B}
        \label{fig:raspberry-cifar10-avg-latency}
    \end{subfigure}
    \hfill
    \begin{subfigure}[b]{0.3\textwidth}
        \centering
        \includegraphics[width=\textwidth]{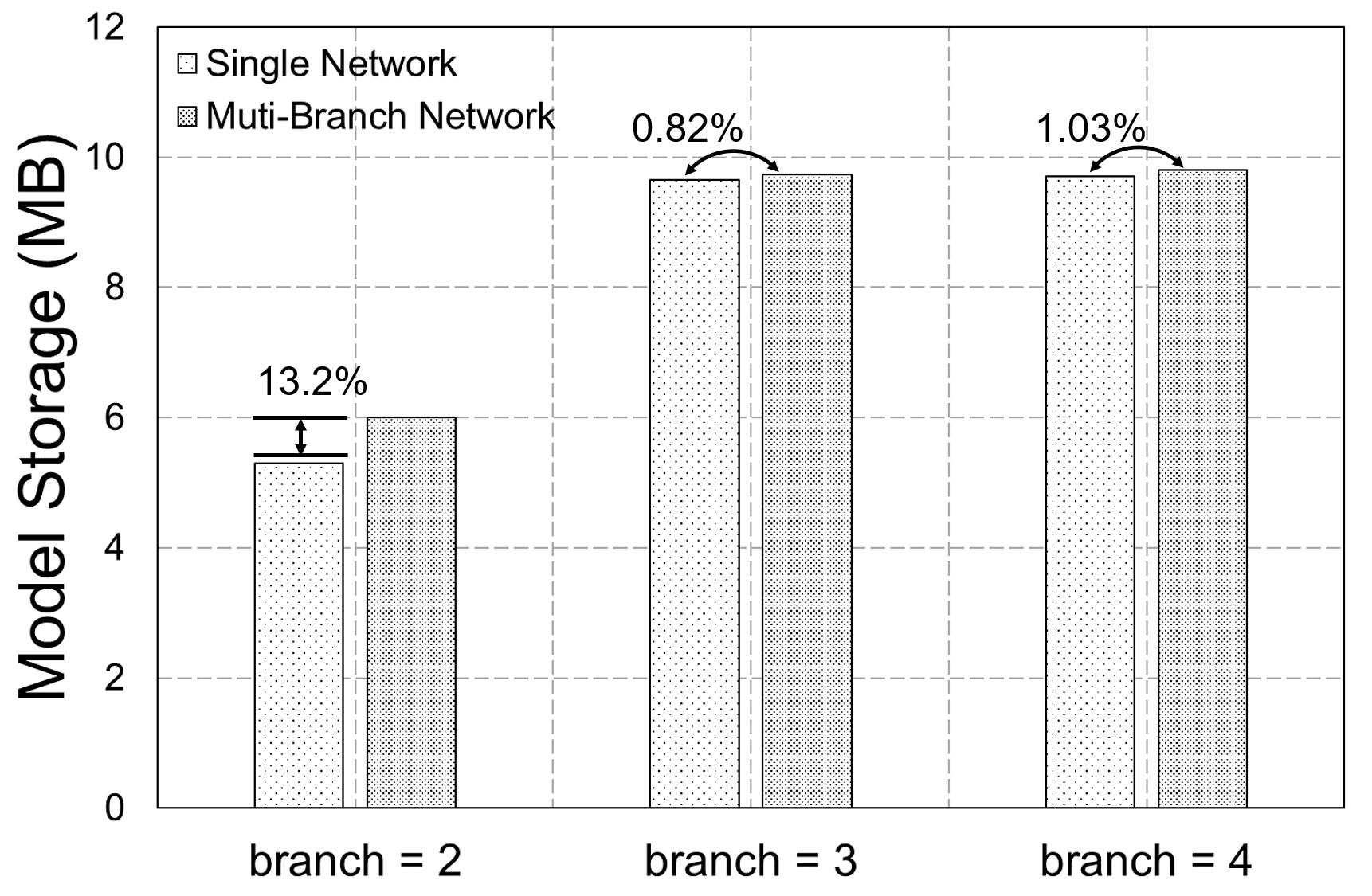}
        \caption{Storage overhead}
        \label{fig:branch_storage}
    \end{subfigure}

    \caption{(a),(b) Average inference latency under dynamic context on mobile devices. (c) Storage overhead with different branches.}
    \label{fig:latency storage}
\end{figure*}

\subsection{Performance Under Dynamic Contexts}
\label{Performance Under Dynamic Contexts}

Evaluating AdaScale's ability to balance accuracy and latency in dynamic scenarios, we deployed it on Jetson Nx and Raspberry Pi 4B. We conducted comparative tests against state-of-the-art lightweight models under dynamic resource conditions. To ensure fair evaluation conditions, a performance control program written in C++ managed resource allocation preemptively in these dynamic contexts.

\begin{figure*}[h]
    \centering
    \begin{subfigure}[b]{0.24\textwidth}
        \includegraphics[width=\textwidth]{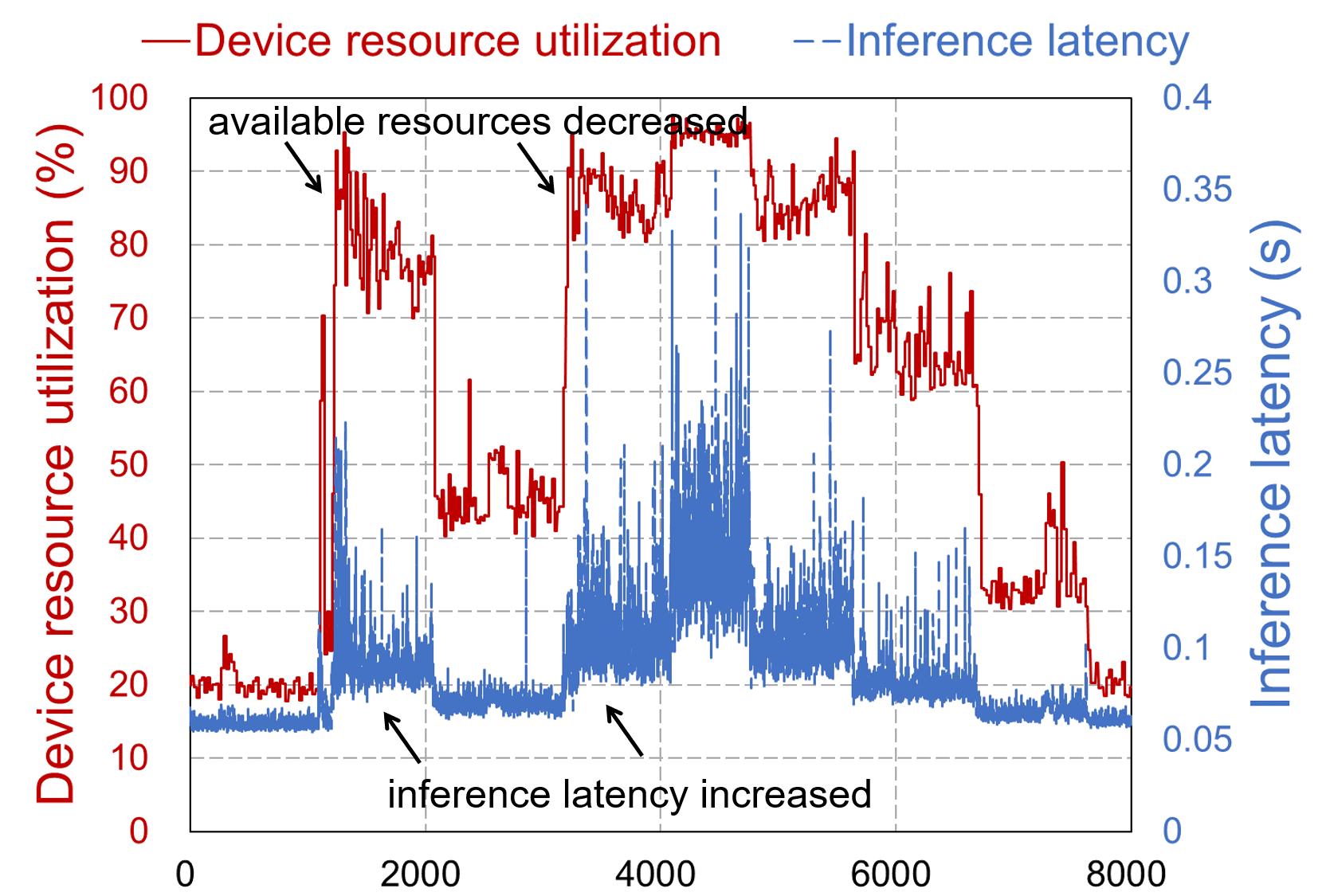}
        \caption{without multi-branch}
        \label{fig:nobranch_latency_resource}
    \end{subfigure}
    \begin{subfigure}[b]{0.24\textwidth}
        \includegraphics[width=\textwidth]{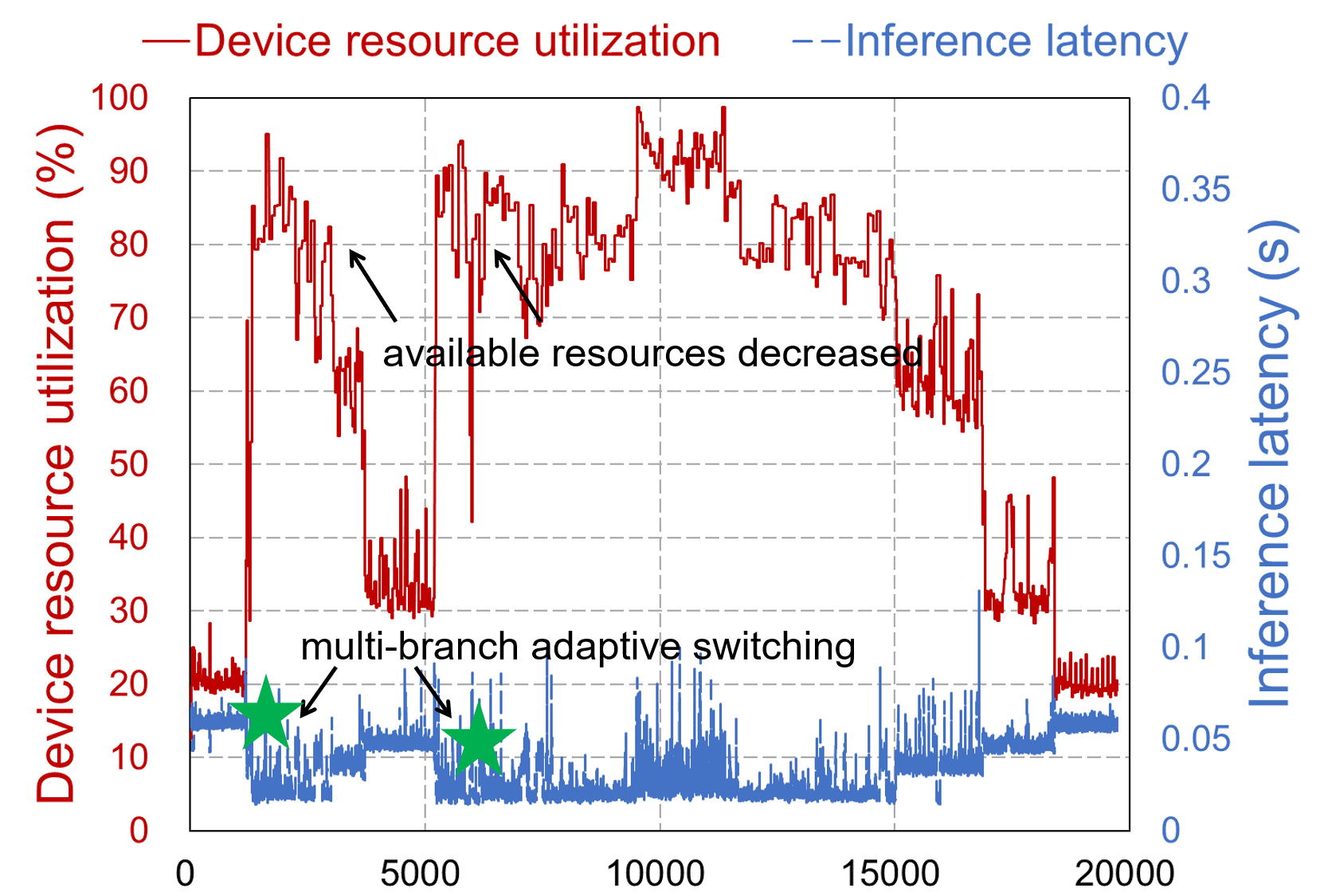}
        \caption{multi-branch adaptation}
        \label{fig:branch_latency_resource}
    \end{subfigure}
    \begin{subfigure}[b]{0.24\textwidth}
        \includegraphics[width=\textwidth]{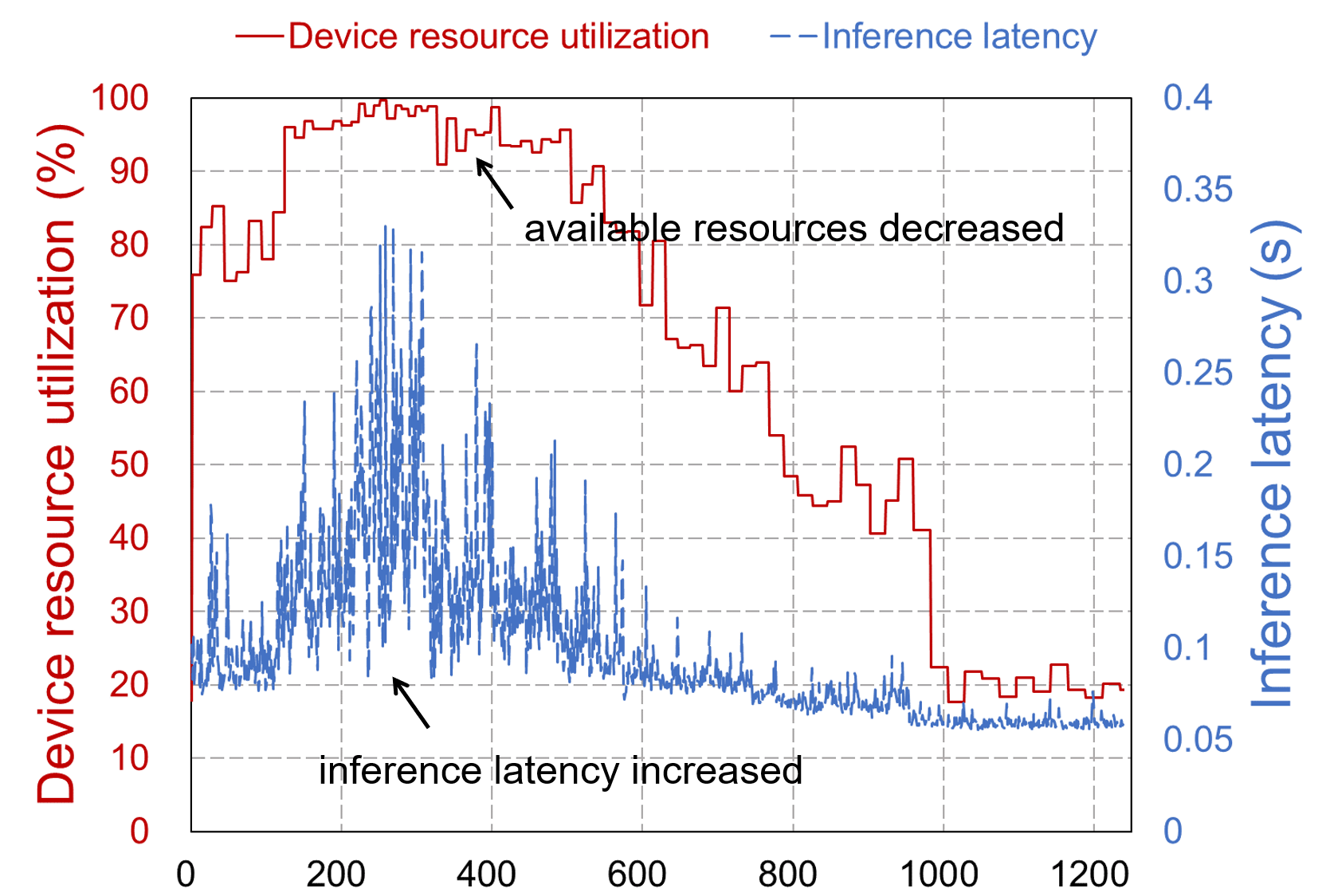}
        \caption{without multi-branch}
        \label{fig:nobranch_latency_resource_2}
    \end{subfigure}
    \begin{subfigure}[b]{0.24\textwidth}
        \includegraphics[width=\textwidth]{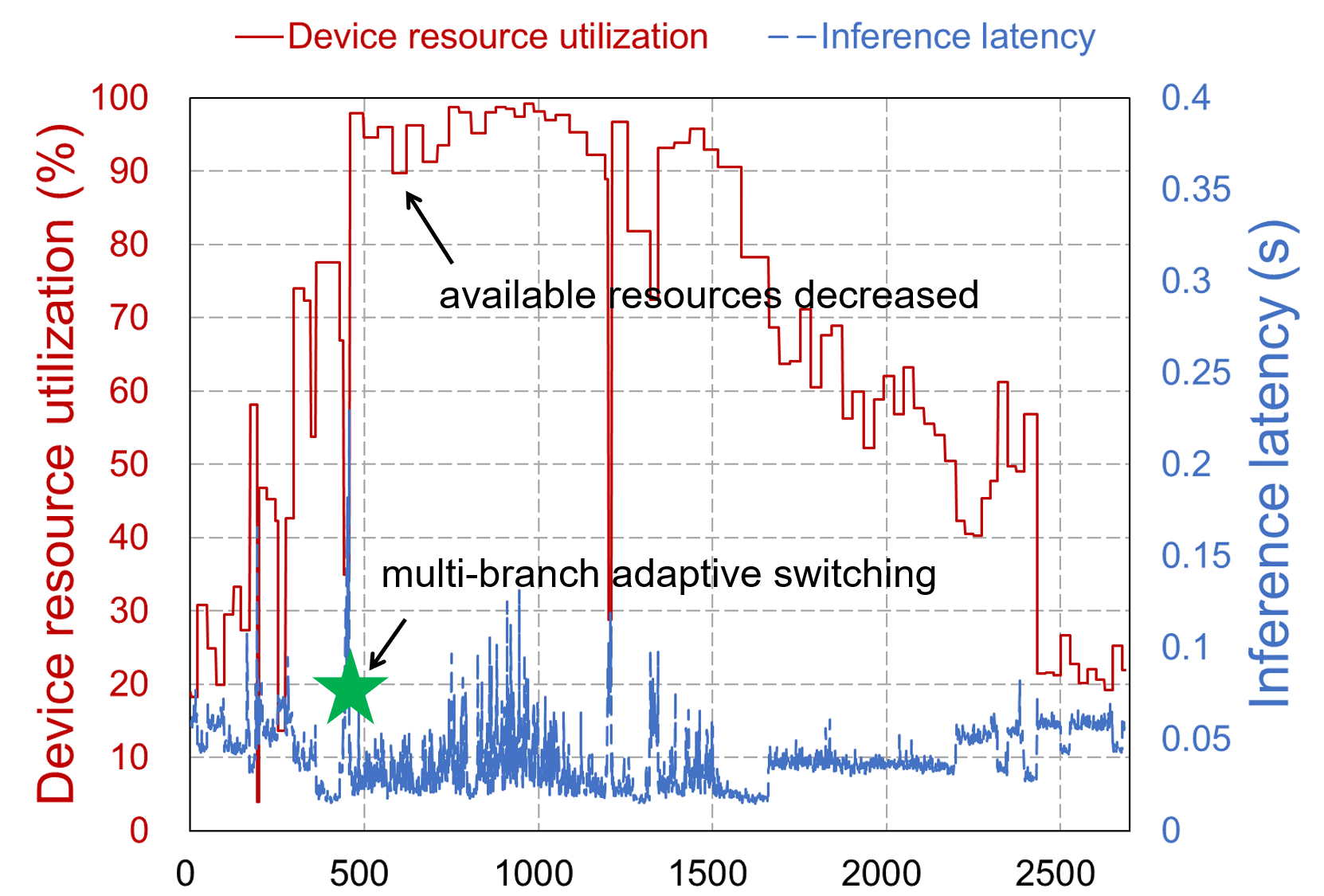}
        \caption{multi-branch adaptation}
        \label{fig:branch_latency_resource_2}
    \end{subfigure}
    
    \caption{AdaScale evaluates device resources and inference latency under different load scenarios. (a) and (b) show resource scenarios under normal device use, (c) and (d) simulate random resource scenarios. (a) and (c) disable the multi-branch adaptive strategy.}
    \label{fig:latency_resource_branch}
\end{figure*}

\begin{figure}[h]
    \centering
    \begin{subfigure}[b]{0.41\textwidth}
        \centering
        \includegraphics[width=\textwidth]{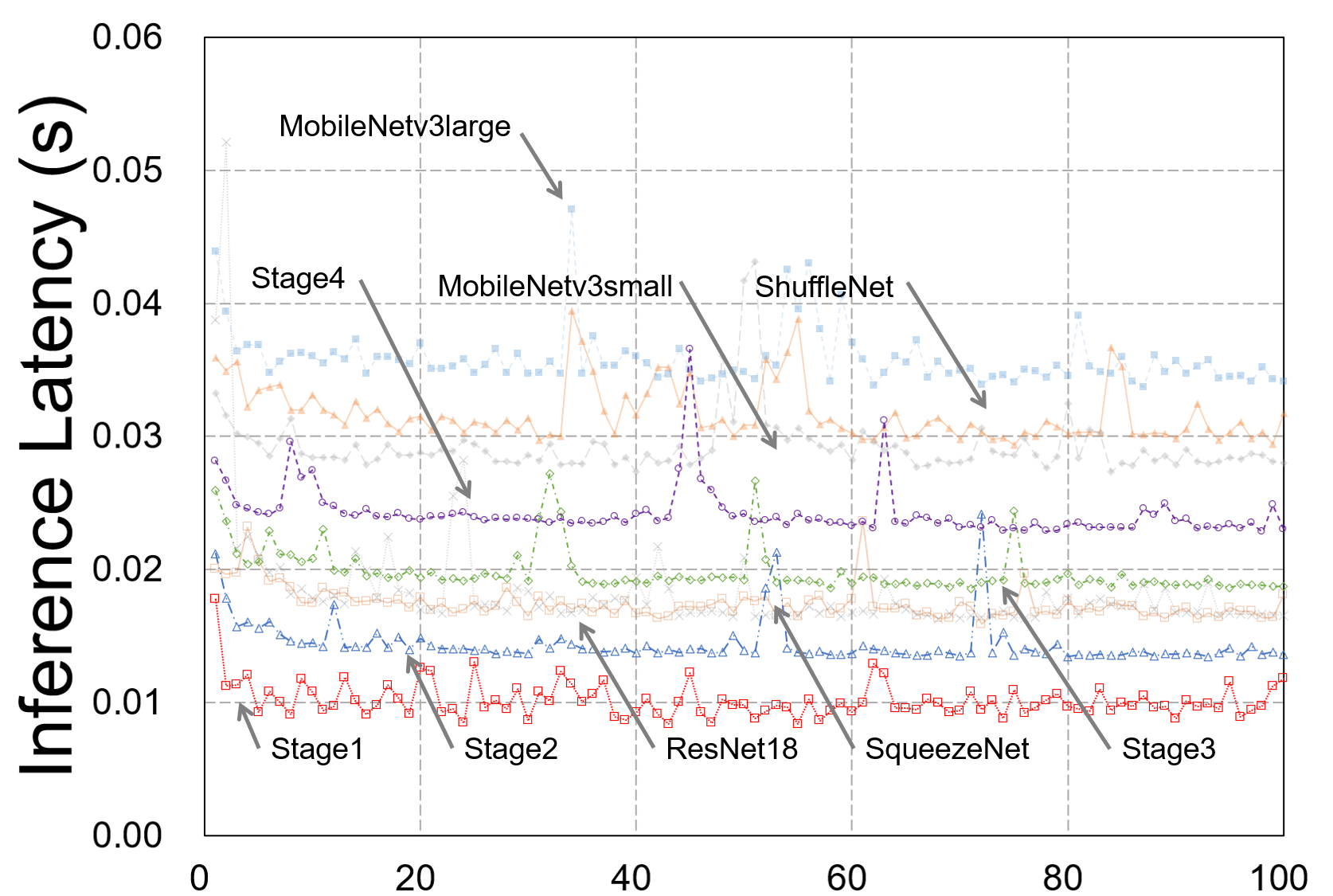}
        \caption{DNNs on Jetson Nx}
        \label{fig:cifar10-latency-static-jetson}
    \end{subfigure}
    
    \begin{subfigure}[b]{0.41\textwidth}
        \centering
        \includegraphics[width=\textwidth]{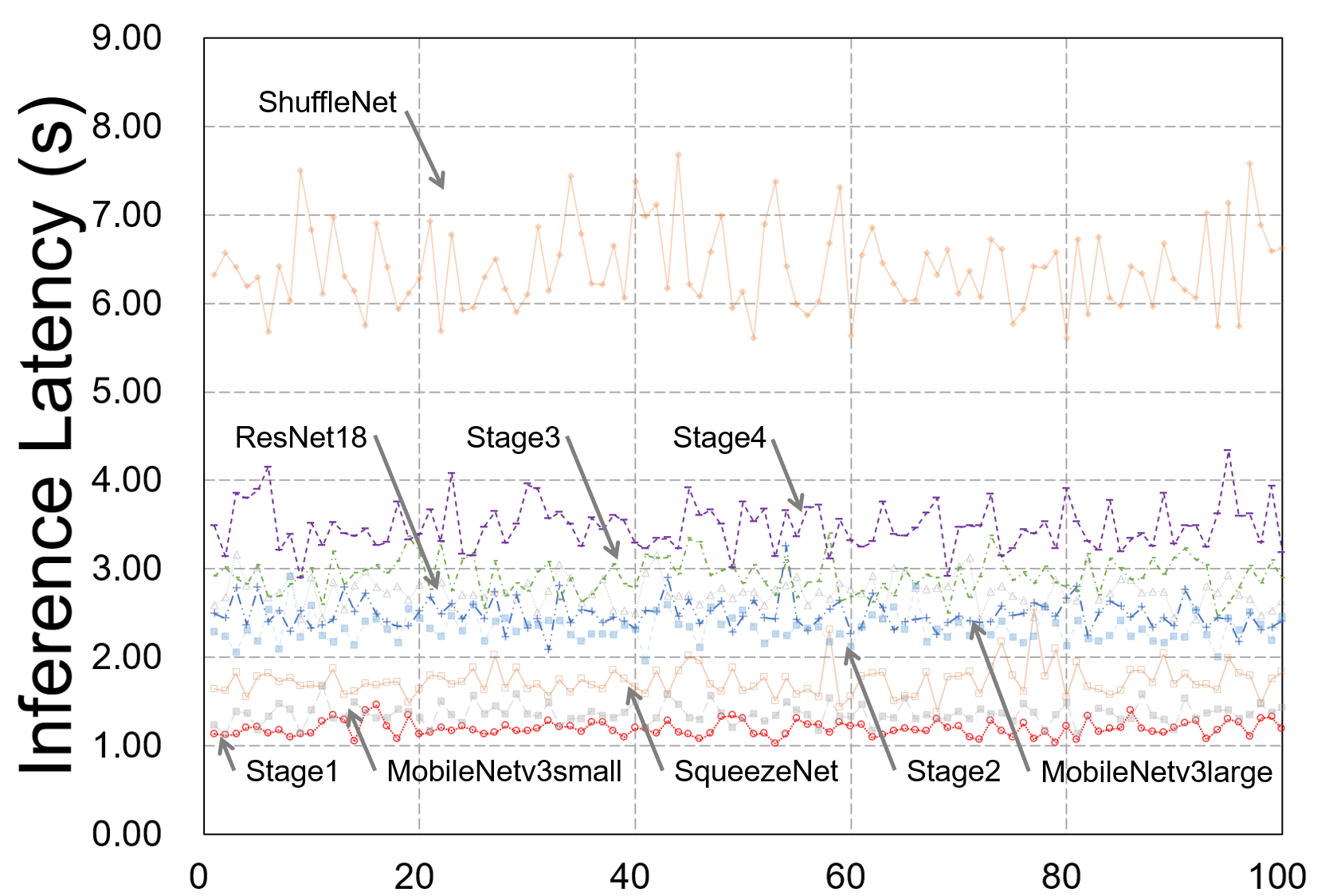}
        \caption{DNNs on Raspberry Pi 4B}
        \label{fig:cifar10-latency-static-raspberry}
    \end{subfigure}

    \caption{Inference latency of different DNNs on mobile devices with resource-constrained context.}
    \label{fig:cifar10-latency-static}
\end{figure}

\textbf{Performance.} As demonstrated in Figure~\ref{fig:dynamic context}, \ref{fig:jetson-cifar10-avg-latency} and \ref{fig:raspberry-cifar10-avg-latency}, AdaScale achieves the lowest inference latency compared to baseline models in all scenarios, with significant improvements under limited resource conditions. AdaScale's optimization strategies enhance resource utilization and reduce inference latency on Jetson Nx and Raspberry Pi 4B devices compared to other advanced lightweight networks. This is achieved by dynamically adjusting the computational load, leading to substantial performance enhancements. Inference speed increases range from 1.14 to 1.85 $\times$ on the Jetson Nx and from 1.71 to 6.29 $\times$ on the Raspberry Pi 4B, with speeds on the Raspberry Pi 4B increasing up to 6.29 $\times$ faster than ShuffleNet. Additionally, AdaScale maintains high model accuracy, up to 86\%, with accuracy losses kept below 4\%, effectively balancing performance and accuracy.


\textbf{Summary:} AdaScale is an effective solution for resource-constrained dynamic environments, enhancing the performance of deep learning models on mobile devices.

\subsection{Performance Under Resource-constrained Contexts}
\label{Performance Under Static Contexts}

We deployed a four-stage elastic scalable network on both Jetson Nx and Raspberry Pi 4B under resource-constrained conditions, comparing its performance with other lightweight networks in the baseline setup.

\textbf{Performance.} As shown in Figure~\ref{fig:cifar10-latency-static}, our multi-stage training method consistently achieves the lowest inference latency on two types of heterogeneous devices, with AdaScale Stage1 consistently outperforming all baselines. On the Jetson Nx and Raspberry Pi 4B, the lowest inference latencies are merely 0.01s and 1s, respectively.. In contrast, models like Mobilenetv3 and Shufflenet show significant performance variability across different devices. This consistency underscores AdaScale's superior adaptability to heterogeneous devices and its ability to effectively balance accuracy and latency. 

\textbf{Summary:} AdaScale offers faster inference speeds under resource-constrained conditions and maintains stable performance across different devices.

\subsection{Performance of Multi-variant Network}
\label{Performance of Multi-variant Network}

We evaluated AdaScale's performance under varying resource conditions on mobile devices by implementing a multithreaded device resource sensing module, a search strategy, and a model performance-directed multi-branch early exit strategy, on Jetson Nx using Python's threading library. This configuration ensured that thread operations did not interfere with each other during runtime. Our tests were conducted under two device load scenarios: random resource changes and modeled user interactions. These tests utilized techniques such as increasing idle read/write operations and streaming web videos to modify the device resource context.

\textbf{Performance.} As shown in Figure~\ref{fig:latency_resource_branch}, introducing a multi-branch early exit strategy significantly reduces inference latency. In scenarios simulating user activity, models without this strategy recorded an average inference latency of 0.0832 seconds. However, models with the multi-branch adaptive strategy achieved a reduced average inference latency of 0.0332 seconds, marking a decrease of 60.10\%. Similarly, in random load scenarios, non-adaptive models exhibited an average latency of 0.09857 seconds, whereas adaptive models decreased it to 0.03924 seconds, a reduction of 60.19\%. Notably, the accuracy drop in these adaptive scenarios did not exceed 5.2\%, as detailed in Tables~\ref{tab:cifar10withmodels} and~\ref{tab:cifar100withmodels}.

\textbf{Summary:} AdaBranch significantly reduces runtime inference latency under varying resource-constrained conditions while maintaining good accuracy performance. AdaScale's each exit branch can be considered a small network, which consistent with the conclusions of recent studies~\cite{wang2024tiny}.

\subsection{Memory Usage of Multi-variant Network}
\label{Memory Usage}

We evaluated the storage overhead of AdaScale by implementing multi-branch networks with two, three, and four branches, alongside their corresponding single-branch networks, on the CIFAR-10 dataset. As shown in Figure~\ref{fig:branch_storage}, the storage overhead from adding branches decreases as the model size increases, falling from 13.2\% with two branches to just 1.03\% with four branches. The overhead for three branches is approximately 1\%, which is negligible. Despite this minimal overhead, the performance improvements achieved through these branches are significant.
accuracy.


\section{RELATED WORK}
\label{RELATED WORK}

 In this section, we discuss the closely related works.


\textbf{Handcrafted DNN Compression.} Applying deep learning-driven intelligent applications to mobile and embedded devices, such as smartphones, IoT devices, and wearables, to enhance various aspects of human life is a promising trend~\cite{liu2020pmc, wen2023adaptivenet}. Significant effort has been invested in manual DNN compression to enable operation on resource-constrained mobile devicess~\cite{disabato2021distributed, lee2020learning}. This involves techniques like quantization~\cite{gholami2022survey}, pruning~\cite{han2015learning}, weight sharing~\cite{han2015deep}, and knowledge distillation~\cite{hinton2015distilling}, all aimed at reducing the size of DNNs~\cite{lee2020learning}. Additionally, efforts have been made to manually create lightweight DNNs, such as MobileNets~\cite{howard2017mobilenets}, CondenseNet~\cite{huang2018condensenet}, and ShuffleNets~\cite{zhang2018shufflenet}. These methods require fine-tuning the width, depth, and structure of DNNs to primarily reduce the number of parameters and storage size. However, few effective compression techniques focus on optimizing application-driven system performance, such as energy efficiency. Recent studies indicate that platform-aware models like SqueezeNet~\cite{iandola2016squeezenet} and SqueezeNext~\cite{gholami2018squeezenext} only reduce parameter size or MAC amount, which does not necessarily lead to reduced energy cost or latency~\cite{choudhary2020comprehensive}. Moreover, these techniques are manually constructed offline and require multi-stage training to ensure accuracy, lacking the capability for dynamic optimization based on specific conditions. In contrast, AdaScale integrates the construction and training of DNNs with runtime device resources. It dynamically optimizes DNNs by considering specific conditions like energy consumption and latency, guiding the optimal specialization.


\textbf{Dynamic-adaptive DNN compression.} Previous research has employed two methods of deployment adaptation: pre-deployment adaptation and post-deployment adaptation~\cite{wen2023adaptivenet}. The former mainly involves cloud-based approaches, where the model is determined in the cloud before deployment to edge devices, such as Neural Architecture Search (NAS)~\cite{liu2021survey}. While effective in finding the optimal model structure based on the target environment, this method is less applicable in dynamically changing mobile environments. The latter is more suitable for complex and varied mobile device scenarios, allowing for precise measurement of model architecture quality in the target environment. For example, AdaptiveNet~\cite{wen2023adaptivenet} generates models for different edge environments, addressing search space and device resource constraints, and LegoDNN~\cite{han2021legodnn} maximizes accuracy under specific resource and latency constraints by training common blocks within DNNs. Some works, like NeuLens~\cite{hou2022neulens} and EdgeCompress~\cite{kong2023edgecompress}, adapt networks based on the complexity of input data, saving resources but often leading to increased training costs due to random generation of descendants. As shown in Figure~\ref{fig_deploy_method}(right), AdaScale employs a novel approach by treating the expansion space as a new set, integrating various lightweight DNN structures (referred to as compression operators) to reduce the search space. It uses multi-stage training before deployment to reduce costs and dynamically selects and combines compression operators based on real-time changes in device resources during runtime. AdaScale utilizes dynamic assessments of device and model performance metrics, such as computational power, energy consumption, and DNN latency, to guide the selection strategy, adapting to various mobile contexts.

\textbf{Runtime Performance Profiler.} Accurately assessing the runtime performance of DNNs on mobile and embedded devices is critical for deploying deep learning effectively~\cite{xu2020latency}. Device parameters such as CPU, GPU, and memory, along with predictive metrics like latency and energy, need to be balanced with DNN parameters including computational complexity, storage, and parameters~\cite{cai2017neuralpower, cai2018proxylessnas, dai2019chamnet}. Mobile devices prioritize runtime latency, energy consumption, and accuracy, and achieving a balance among these is crucial to leveraging the capabilities of both the device and the DNN. Current efforts include Nn-meter, which predicts the runtime latency of DNNs on NVIDIA GPUs using operator fusion~\cite{zhang2021nn}; AdaEnlight, which uses an energy predictor to manage the dynamic energy consumption of DNNs on devices with limited battery capacity~\cite{liu2023adaenlight}; and GPUNet, which uses latency budgets to guide model adjustments for improved accuracy~\cite{wang2022searching,liu2018demand}. These efforts enhance deep learning deployment on mobile edge devices. However, they often fail to consider the dynamically changing contexts of mobile and embedded devices, where DNN runtime latency and energy consumption can vary with changes in device load. In contrast, AdaScale is designed to adapt to these changes by employing a multi-branch elastic network that selects appropriate branches for termination based on resource availability. It also integrates intrinsic device and DNN performance metrics to build an energy and latency predictor, accurately forecasting the diverse contextual demands of device runtime.
\section{CONCLUSION}
\label{CONCLUSION}

In this paper, we introduce AdaScale, an elastic inference framework designed to automatically adapt deep models to dynamic operational contexts. 
AdaScale is engineered to streamline DNN creation using a self-evolutionary model, effectively minimizing the search space and enhancing the quality of the resulting networks through a strategic combination of diverse compression operators. 
Key features of AdaScale include a multi-branch early exit structure that facilitates dynamic resource adaptation and a multi-stage training mode that optimizes network performance while simultaneously reducing training expenditures.
Moreover, AdaScale incorporates advanced performance profiling and resource awareness capabilities, which are crucial for initiating timely and accurate model adaptations. 
This ensures that the framework takes full advantage of real-time device capabilities, significantly boosting adaptability. 
In future work, we aim to explore the efficient generation and deployment of diverse architectural models. Our research will facilitate the deployment of more advanced large-scale models on intelligent IoT devices, thus enhancing the quality of services provided to users.


\bibliographystyle{IEEEtran}
\bibliography{IEEEabrv,adabranch}


\end{document}